\documentclass[11pt]{article}

\usepackage[margin=1in]{geometry}
\usepackage[T1]{fontenc}
\usepackage{lmodern}
\usepackage{amsmath,amssymb}
\usepackage{booktabs}
\usepackage{microtype}
\usepackage{graphicx}
\usepackage[hidelinks]{hyperref}
\usepackage{url}
\usepackage{xcolor}
\usepackage{enumitem}
\usepackage{authblk}
\usepackage{tikz}
\usepackage{longtable}
\usetikzlibrary{positioning, arrows.meta, shapes.geometric, fit, calc}

\title{ChessMimic: Per-Rating Transformer Models for Human Move, Clock, and Outcome Prediction in Online Blitz Chess}

\author{Thomas Johnson\thanks{\texttt{tjohnson@nascent.xyz}. Live demo: \url{https://1e4.ai}. Code: \url{https://github.com/thomasj02/1e4_ai}.}}
\affil{Nascent}

\begin{document}
\maketitle

\begin{abstract}
We present \textsc{ChessMimic}, a system of three small encoder-only
transformers -- for move, thinking-time, and outcome prediction --
conditioned on the position, recent move history, player rating, and
clock state. We fit a separate instance of each model per 100-Elo
rating band, trading parameter efficiency for sharper per-skill
calibration. On a held-out month-wide slice of Lichess Rated Blitz games
\textsc{ChessMimic}'s human move prediction accuracy outperforms Maia-2 
in every Elo band. Compared to Maia-3, our 9M parameter model's accuracy
sits between Maia-3-5M and Maia-3-23M without the additional complexity of
Geometric Attention Bias.

In addition to the move matching model, we also train a game outcome model
that conditions not only on the position, but also player ratings,
time control, and remaining clock times. The outcome model achieves an AUC
of $0.78$ out of sample, beating Maia-2 as well as logistic regressions based
on material, ratings, and clock time.

Finally, we train a clock model that predicts human thinking times. The
clock model provides a usable but non-SOTA per-ply think-time signal
under ALLIE-style filters (Pearson $r=0.41$, Spearman $\rho=0.50$,
MAE $4.10$\,s, against ALLIE's reported $r=0.70$), with the residual
gap concentrated in per-position bucket sharpness rather than
bucket-marginal calibration.

A public demo is at \href{https://1e4.ai}{1e4.ai} and we release
code, per-band weights, and the C++ data-filter pipeline code in GitHub.
\end{abstract}

\section{Introduction}
\label{sec:intro}

Modern chess engines such as Stockfish~\cite{stockfish_nnue} and
AlphaZero~\cite{silver2018alphazero} are superhuman in their ability to play chess, 
but their strength makes them frequently unhelpful for chess practice or recreation.
As the chess writer Tim Krabb\'e said of a perfect tablebase endgame: ``They are not human; a
grandmaster does not understand them any better than someone who has learned chess yesterday.''~\cite{krabbe_endgame}
To address this strength difference, engines can be forced to play at lower levels by injecting 
random moves or extremely short calculation depth. This reduces the engine's strength, but 
still makes moves that are qualitatively different from humans at an equivalent rating.
A separate research thread therefore studies
\emph{human-aligned} chess play -- engines whose objective is to imitate
human move choices, rather than to win. The \textsc{Maia} family
\cite{mcilroyyoung2020maia,mcilroyyoung2022individual,tang2024maia2,tang2025maia4all}
established the modern framing: condition on the player's rating, train on
millions of online human games, and evaluate top-$k$ accuracy against the human-played
move at various Elo ranges.

Two limitations of pure move-prediction systems become apparent once they are
deployed as opponents in a real application:

\begin{enumerate}[leftmargin=*,topsep=2pt,itemsep=2pt]
  \item \textbf{Clock realism.} Blitz games are decided as much by time pressure
        as by move quality. A bot that returns instantly, or always uses the
        same time per move, feels unrealistic and does not allow for time-related
	strategies like complicating the position to cause your opponent to use
	more time. Bots that move unrealistically fast may also degrade the perceived
	realism of blitz games, since human play involves meaningful
	opponent thinking intervals and time allocation patterns that prior
	work has shown to be behaviorally informative~\cite{russek2025value,rheude2021ctm}.
  \item \textbf{Outcome forecasting.} A win / loss / draw outcome head without clock
        conditioning can capture material balance but is poorly suited to the
        games we actually want to model: human-vs-human blitz, where
        flagging, low-time blunders, and stalemate-from-winning positions are
        common.
\end{enumerate}

\noindent\textsc{ChessMimic} addresses both concerns directly. In addition
to move models trained on human games, we also train a thinking time model
that can be used to implement synthetic delays that mimic human thinking time,
and an outcome (i.e. win/loss/draw) model that takes into account the position,
player ratings, and players' clock times.
A public demo at \href{https://1e4.ai}{1e4.ai} lets users play against 
ChessMimic opponents calibrated to their rating, with human-like time usage
and a UI-visible win-probability bar. 
Architecturally ChessMimic is a rating-specific set of three simple, 
independent small encoder-only transformers 
($\sim$256-dim embeddings, 8 layers, 9M parameters): a move model, a clock model, and a
winner model. These share a common input tokenization and a common encoder
template, and each set is trained independently on a 100-point-wide rating band
of Lichess blitz games. The deployed system contains 14 rating bands $\times$ 3 task
models $=$ 42 distinct checkpoints.

The contributions of this work are:

\begin{itemize}[leftmargin=*,topsep=2pt,itemsep=2pt]
\item A simple, deployable transformer recipe -- shared tokenization and a
  shared encoder template, instantiated as three independently trained
  task-specific models -- that covers move, clock, and outcome prediction
  with per-model checkpoints small enough to serve on a single commodity
  CPU.
\item A direct head-to-head benchmark of \textsc{ChessMimic} against the
  Maia-2 blitz model on a month-wide held-out slice of Lichess April 2026
  Rated Blitz games -- one randomly chosen non-opening position from each
  of 4.79 million games -- including both move accuracy and outcome
  calibration.
\item A demonstration that a separate 3-class winner model (W/D/L) with
  clock conditioning produces a substantially better-calibrated
  expected-score signal for human blitz than the scalar value-head
  alternatives we evaluate, while reusing the same encoder template and
  rating conditioning as the move model.
\end{itemize}

\section{Related Work}
\label{sec:related}

\paragraph{Engines vs.\ human models.}
The dominant strand of computer chess research targets the strongest play
possible. AlphaZero~\cite{silver2018alphazero} demonstrated that self-play
reinforcement learning over a residual network with Monte-Carlo tree search
could reach grandmaster strength tabula rasa, and Stockfish has integrated
neural network evaluation (NNUE)~\cite{stockfish_nnue} since 2020.
\textsc{ChessMimic}'s objective is the opposite: imitate human play at a
specified rating, including its characteristic blunders and time-management
patterns.

\paragraph{Move imitation: the Maia family.}
McIlroy-Young et al.~\cite{mcilroyyoung2020maia} introduced Maia, a
fixed-depth AlphaZero variant trained on rating-bucketed human games which
matches human moves at 46-52\% top-1 accuracy depending on skill level. A
personalized follow-up \cite{mcilroyyoung2022individual} fine-tunes Maia on
individual players' game histories, and a stylometry study
\cite{mcilroyyoung2021stylometry} showed that 100 games suffice to identify
players among thousands. Tang et al.~\cite{tang2024maia2} introduced Maia-2,
a single skill-conditioned model with a learned skill-aware attention
mechanism that unifies the rating-bucketed Maias and beats them by roughly
two percentage points top-1. Maia4All~\cite{tang2025maia4all} pushes the
personalization frontier, requiring only 20 games per player to fit an
individual embedding. Most recently, Monroe et al.~\cite{monroe2026chessformer}
introduce Chessformer (also referred to as Maia-3 in its human-modeling
configuration), a unified architecture that combines Maia-style human
modeling with a Leela-BT4-derived engine variant. Its central novelty is a
chess-specialized positional encoding called Geometric Attention Bias (GAB):
a dynamic generator that emits per-head additive bias tensors of shape
$h\times 64 \times 64$ over square pairs, conditioned on the position, which
are added to the attention logits before softmax. The motivation is that
chess geometry is piece-type- and state-dependent (e.g.\ locked pawn chains
weaken bishop diagonals), so a static positional encoding mapped onto a
1D linearization of the board is misaligned with the action space.
Chessformer reports 57.1\% top-1 at 79M parameters and 56.6\% at 23M
parameters on the ALLIE blitz test set, leaving Maia-2 at 52.0\% on the same
slice. Our work occupies a complementary spot in this design space: we
deliberately retain Maia-1's per-rating ensemble structure, which is easy to
reason about and to serve, while extending the prediction surface to clocks
and outcomes. The positional-encoding choice in \textsc{ChessMimic} is the
fully learned 1D embedding of Ruoss et al.~\cite{ruoss2024searchless}, not
GAB; closing that architectural gap is a clear follow-up direction.

\paragraph{Time-management and outcome modeling.}
Rheude~\cite{rheude2021ctm} trained MLPs and CNNs to predict per-ply
thinking time from board features, and showed that human time usage is more
than a simple decay of the remaining clock. Omori and
Tadepalli~\cite{omori2024rating} estimate a player's rating from the
\emph{joint} sequence of moves and clock readings using a CNN-LSTM, achieving
$\sim$182 Elo points MAE; this is direct evidence that clock usage carries
identity- and skill-level information that pure-board models cannot
recover. Russek et al.~\cite{russek2025value} analyzed 12M Lichess games and
showed that humans spend more time thinking in positions where additional
computation is rational, with stronger players exhibiting a tighter
relationship. The recent ALLIE system~\cite{zhang2025allie} treats game
transcripts including clock readings as a single autoregressive sequence,
producing a model whose ``pondering'' decisions correlate strongly with
human thinking times ($r\approx0.70$) and which also learns a value head
over game outcomes used as an MCTS budget signal. Outcome/value modeling for
\emph{human} games -- as opposed to engine evaluation -- is comparatively
under-explored; \textsc{ChessMimic}'s separately-trained winner model is our
contribution to that gap.

\paragraph{Transformer chess.}
The transformer architecture~\cite{vaswani2017attention} has only recently
become a strong backbone for chess. Ruoss et al.~\cite{ruoss2024searchless}
trained encoder-only transformers up to 270M parameters on $10^{10}$
Stockfish-annotated positions, reaching a 2895 Lichess blitz rating without
search. Their FEN tokenizer and \texttt{bagz} record format have proved
broadly useful; we reuse both in \textsc{ChessMimic}'s data pipeline.

\paragraph{Evaluation.}
For move prediction we report top-$k$ accuracy against the actually played
move, the standard metric in
\cite{mcilroyyoung2020maia,tang2024maia2,zhang2025allie}. For outcome
prediction we use the Brier score~\cite{brier1950score}, a strictly proper
scoring rule that rewards both reliability and resolution, together with
log-loss and ROC AUC for completeness. For our self-play rating experiments
we use the Glicko-2 system~\cite{glickman2012glicko2}.

\section{Data}
\label{sec:data}

Training data is built from monthly Lichess PGN dumps
\cite{lichess_database}. A C++ pipeline parses \texttt{[\%clk]}
annotations, computes per-ply thinking times, and writes records to
the \texttt{bagz} format \cite{ruoss2024searchless}. The per-band
rating filter applied at training time differs by task model, and
also differs from the per-band routing used at serving time
(Table~\ref{tab:band_routing}). The resulting train/serve
distribution shift -- training sees a stricter slice of the rating
distribution than serving admits -- is a small source of mismatch we
have not separately quantified.

\begin{table}[h]
\centering\small
\begin{tabular}{l l l}
\toprule
Model & Training-time band filter & Serving-time routing \\
\midrule
Move    & both white and black Elo in $[\text{lo},\text{hi}]$ & side-to-move's Elo \\
Clock   & both white and black Elo in $[\text{lo},\text{hi}]$ & side-to-move's Elo \\
Winner  & $\lfloor (\text{white\_elo}+\text{black\_elo})/2 \rfloor \in [\text{lo},\text{hi}]$ & average of the two players' Elos \\
\bottomrule
\end{tabular}
\caption{Per-band routing for the three task models. Move and clock
  training require both players to be in the band; winner training
  requires their integer average to be in the band, matching the
  average-rating serving route. The training filter is therefore
  strictly stronger than the serving router for move and clock (only
  the side-to-move is routed at serve time, so the opponent can be
  out-of-band), and equal in form for winner (both train and serve
  key on the average).}
\label{tab:band_routing}
\end{table}

Two practical filters significantly improve training efficiency:

\begin{enumerate}[leftmargin=*,topsep=2pt,itemsep=2pt]
  \item \textbf{Common-position skipping.} Positions whose
        \texttt{(FEN, last-12-moves)} key has been seen more than 1000 times
        on the training shard are skipped during the data build, so the
        network does not waste capacity memorizing book moves. Their
        empirical move distributions are retained in a per-band database
        that is consulted at serve time (\S\ref{sec:deployment}).
  \item \textbf{Blitz filter.} We keep only games Lichess categorizes as
        blitz (PGN \texttt{Event}~$=$~``Rated Blitz game''), using
        Lichess's own speed classification rather than a numeric
        time-control threshold of our own. Bullet, rapid, and classical
        games have visibly different time-usage distributions and are out
        of scope for this release. The \texttt{TimeControl} tag is still
        parsed, but only to extract the base time and increment as model
        features, not as a filter.
\end{enumerate}

\paragraph{Train / validation / test split.} Training uses the 12
consecutive Lichess monthly dumps from 2024-09 through 2025-08 inclusive.
Validation uses a $42{,}060$-game slice of the 2025-09 dump. The test set
is the 2026-04 dump -- seven months after the validation month and
disjoint from both training and validation, and used for neither training
nor for building any common moves database, so all reported numbers are
genuinely held out. (This also makes the 2022 ALLIE blitz test slice,
\S\ref{sec:chessformer_compare}, clean held-out data for
\textsc{ChessMimic}, since no 2022 month enters training.) For the
held-out test slice we take \emph{all} $37{,}580{,}604$ Rated Blitz games
in the April 2026 dump, filter inline to \texttt{Rated Blitz}, exclude
games in which either player has the Lichess \texttt{BOT} title so the
comparison is strictly human-vs-human (bot games are concentrated at the
top of the rating range, $\approx$2.3\% of pre-filter 2200+ positions
vs $\approx$0.3\% overall), skip the first 8 plies of each game (opening
book), and drop games shorter than 20 plies. From each surviving game
we keep exactly \emph{one} uniformly random eligible (non-opening) ply,
with the choice seeded by the Lichess game id so the sample is
reproducible and independent of processing order. One ply per game removes within-game correlation and makes the
closed-form confidence intervals in \S\ref{sec:experiments} much
better-behaved; residual between-game clustering (repeated players,
opening repertoires, same-day tournaments) is a real but smaller
source of optimism. For the headline benchmark we score a uniformly random
$5$-million-game subset, yielding $n=4{,}775{,}033$ held-out positions in
the natural Lichess rating distribution across all 14 \textsc{ChessMimic}
bands (per-band $n$ ranges from $\approx$179k in the 2100--2200 band to
$\approx$490k in the 1600--1700 band, with $\approx$276k in the sub-1000
band; every band has $\gg$100k positions). The more expensive
counterfactual studies in \S\ref{sec:clock_studies} (Stockfish blunder
labeling, clock and move-distribution sweeps, thinking-time correlation)
instead use a fixed $100{,}000$-position sub-sample drawn from the same
set and stratified to $7{,}143$ positions per band, so that even the
smallest bands are well represented for per-band breakdowns.

\section{Model}
\label{sec:method}

\subsection{Inputs and Tokenization}

All three tasks share the same per-position input. The board is a FEN
string tokenized into a length-78 sequence using the searchless-chess
tokenizer~\cite{ruoss2024searchless}: 64 squares (with run-length expansion
of empty squares to single dots), a side-to-move token, four castling
rights, two en-passant characters, and the halfmove clock and fullmove
number, each tokenized to a fixed-width field, plus a trailing CLS token.
Vocabulary size is 33 (31 chess characters, CLS, PAD). Because the fullmove
number is part of the tokenized FEN, the model already sees how far into the
game the position is; we do not add any separate move-count input. The
recent-move input is the last 12 plies in UCI form, each mapped to one of
$|\mathcal{A}|$ legal-move action ids and left-padded with a dedicated PAD
token; $\mathcal{A}$ enumerates all combinatorially possible chess
moves -- 1792 non-promotion plus 176 promotion entries, $|\mathcal{A}|=1968$
total. The rating is the to-move player's Elo, standardized to zero mean and
unit variance using statistics fit on the training shard. The clock state
is, for the move model, the to-move player's remaining time, log-transformed
and standardized; for the clock model, a 3-vector $[\log(t_\text{self}+1),
\log(t_\text{opp}+1), \log(\text{inc}+1)]$, each component standardized; and
for the winner model, a single 5-vector $[r_\text{white}, r_\text{black},
t_\text{white}, t_\text{black}, \text{inc}]$ (ratings standardized, times
log-scaled) that replaces the separate rating and clock embeddings, so that
all side-information is carried by a single token.

\subsection{Architecture}

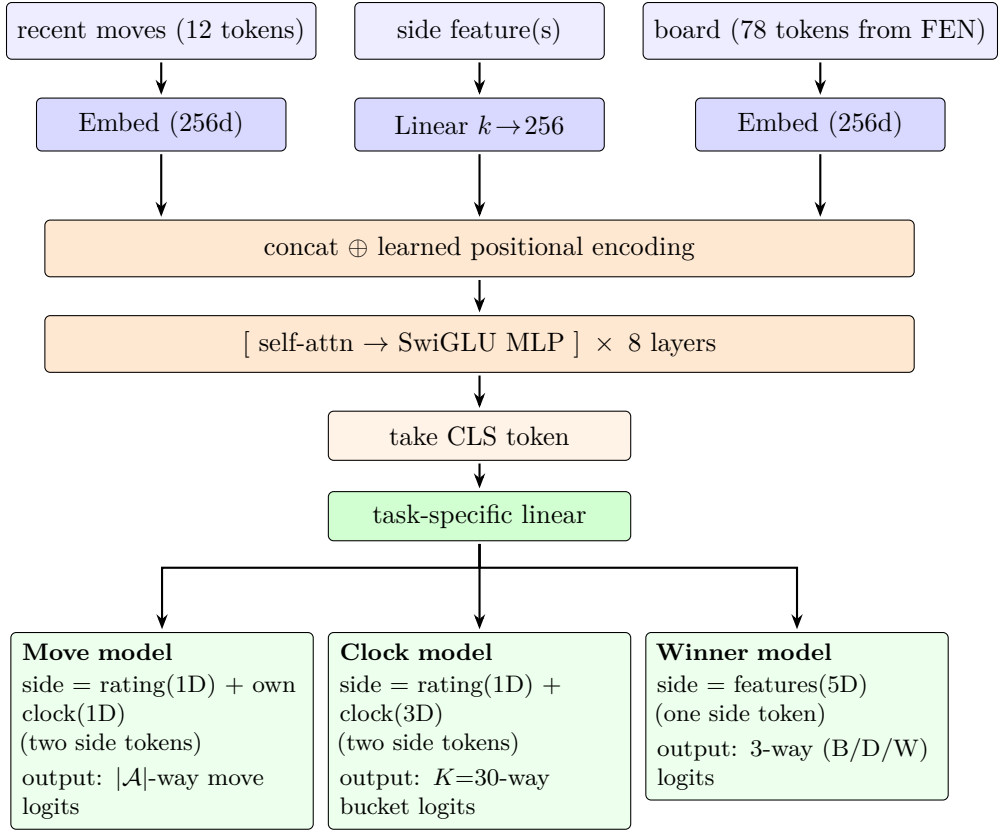
\begin{figure}[t]
\centering
\begin{tikzpicture}[
    font=\small,
    >={Stealth[length=2mm]},
    every node/.style={align=center, inner sep=4pt},
    input/.style={draw, rounded corners=2pt, fill=blue!7, minimum width=3.3cm, minimum height=0.75cm},
    embed/.style={draw, rounded corners=2pt, fill=blue!15, minimum width=3.3cm, minimum height=0.7cm},
    trunk/.style={draw, rounded corners=2pt, fill=orange!18, minimum width=11.5cm, minimum height=0.75cm},
    midbox/.style={draw, rounded corners=2pt, fill=orange!10, minimum width=4.0cm, minimum height=0.65cm},
    outhead/.style={draw, rounded corners=2pt, fill=green!18, minimum width=4.0cm, minimum height=0.65cm},
    variant/.style={draw, rounded corners=2pt, fill=green!7, minimum width=4.0cm, minimum height=1.7cm, align=left, text width=3.7cm, font=\footnotesize},
    arr/.style={->, thick, shorten >=1pt, shorten <=1pt},
  ]

  \node[input] (recent) {recent moves (12 tokens)};
  \node[input, right=0.5cm of recent] (side)   {side feature(s)};
  \node[input, right=0.5cm of side]   (board)  {board (78 tokens from FEN)};

  \node[embed, below=0.5cm of recent] (eRecent) {Embed (256d)};
  \node[embed, below=0.5cm of side]   (eSide)   {Linear $k\!\to\!256$};
  \node[embed, below=0.5cm of board]  (eBoard)  {Embed (256d)};

  \draw[arr] (recent) -- (eRecent);
  \draw[arr] (side)   -- (eSide);
  \draw[arr] (board)  -- (eBoard);

  \node[trunk, below=0.9cm of eSide] (concat) {concat $\oplus$ learned positional encoding};

  \draw[arr] (eRecent.south) -- ++(0,-0.35) -| (concat.north -| eRecent);
  \draw[arr] (eSide.south)   -- (concat.north -| eSide);
  \draw[arr] (eBoard.south)  -- ++(0,-0.35) -| (concat.north -| eBoard);

  \node[trunk, below=0.5cm of concat] (trans)
    {$[$ self-attn $\to$ SwiGLU MLP $]\;\times\;8$ layers};
  \draw[arr] (concat) -- (trans);

  \node[midbox, below=0.5cm of trans] (cls) {take CLS token};
  \draw[arr] (trans) -- (cls);

  \node[outhead, below=0.4cm of cls] (head) {task-specific linear};
  \draw[arr] (cls) -- (head);

  \node[variant, below=1.2cm of head, xshift=-4.2cm] (vmove)
    {\textbf{Move model}\\[1pt]
     side $=$ rating(1D) $+$ own clock(1D)\\
     (two side tokens)\\[2pt]
     output: $|\mathcal{A}|$-way move logits};

  \node[variant, below=1.2cm of head] (vclock)
    {\textbf{Clock model}\\[1pt]
     side $=$ rating(1D) $+$ clock(3D)\\
     (two side tokens)\\[2pt]
     output: $K{=}30$-way bucket logits};

  \node[variant, below=1.2cm of head, xshift=4.2cm] (vwinner)
    {\textbf{Winner model}\\[1pt]
     side $=$ features(5D)\\
     (one side token)\\[2pt]
     output: 3-way (B/D/W) logits};

  \draw[arr] (head.south) -- ++(0,-0.35) -| (vmove.north);
  \draw[arr] (head.south) -- (vclock.north);
  \draw[arr] (head.south) -- ++(0,-0.35) -| (vwinner.north);

\end{tikzpicture}
\caption{\textsc{ChessMimic} encoder template. We instantiate this template
  three times -- as a move model, a clock model, and a winner model. Each
  instance is trained independently on its own bagz dataset with a Brier
  loss, and the three instances have distinct weights at inference time.
  One move model is trained from scratch to convergence; every other move
  model, and all clock and winner models, is fine-tuned from that root
  move checkpoint (transformer blocks transferred unchanged, the
  side-feature embedding widened to the new arity, and the output
  projection re-initialized) before being fine-tuned end-to-end. See
  \S\ref{sec:method} for the exact cascade.}
\label{fig:arch}
\end{figure}

Each of the three task models is an encoder-only transformer with the
structure shown in Figure~\ref{fig:arch}. The three models share this
template but are trained independently and have their own weights at
inference time; they differ in the dimensionality and packing of their
side-feature tokens, in the output projection, and in the dataset they are 
trained on.

After embedding board and recent-move tokens to a 256-dimensional space,
the scalar/vector side features (rating, clock, etc.) are projected through
small linear layers and inserted as additional ``tokens'' immediately before
the board tokens. A learned positional encoding of shape
$(L_{\text{recent}}+L_{\text{side}}+L_{\text{board}}) \times 256$ is added,
where $L_{\text{recent}}=12$, $L_{\text{board}}=78$, and $L_{\text{side}}\in
\{1,2\}$ depending on the model (1 for the winner model, which packs all
side information into a single 5-D vector; 2 for the move and clock models,
which use separate rating and clock embeddings). The encoder consists of 8
identical residual blocks alternating a pre-norm multi-head self-attention
layer (8 heads, no relative bias) with a SwiGLU MLP~\cite{shazeer2020glu}
of width $4\times256=1024$. The CLS token at the end of the board sequence
is read out and projected by the task-specific output layer. We
deliberately avoid bias terms in the MLP and use SiLU-gated linear units,
following modern transformer practice.

The three models differ only in their inputs (as above), their final
output projection, and the dataset they consume:

\paragraph{Move model.} Output: linear $256\to|\mathcal{A}|$. Logits over
illegal moves are masked to $-\infty$ before softmax, and the loss is the
unnormalized masked Brier
$L = \sum_{a\in\mathcal{L}(s)} (p_a - y_a)^2$, summed over legal moves only
($\mathcal{L}(s)$ is the set of legal moves in state $s$ and
$y_a\in\{0,1\}$ is the one-hot played-move target). We use Brier instead
of cross-entropy~\cite{brier1950score} because our target is inherently
probabilistic: even the same player in the same position may prefer
one move sometimes and another move at a different time, depending on
their mood, their opponent, and other unmodeled factors.

\paragraph{Clock model.} Output: linear $256 \to K$ where $K=30$ is the
number of thinking-time buckets. Bucket boundaries are fit on the training
shard using a hybrid linear-quantile scheme: 1-second buckets for $t \in
[0, 27]$ seconds (covering $\approx$86\% of blitz moves), then progressively
wider buckets up to a half-open final bucket at 40+ seconds. The loss is
the same masked-Brier form
$\sum_{b} (p_b - y_b)^2$ over the $K$ buckets (buckets that contain thinking
time larger than what the player has on their clock are masked).

\paragraph{Winner model.} Output: linear $256 \to 3$, producing logits for
black-wins / draw / white-wins. The loss is the 3-class Brier
$\sum_{c\in\{B,D,W\}} (p_c - y_c)^2$ over the full simplex; there is no
masking. At inference time, for the purposes of comparison with Maia-2's
scalar value head, the 3-class output is collapsed to the side-to-move's
\emph{expected score} $P(\text{side wins}) + 0.5\cdot P(\text{draw})$
(targets $y\in\{1.0,\,0.5,\,0.0\}$ for win/draw/loss). The Brier and
log-loss numbers reported in \S\ref{sec:experiments} are computed
against this scalar target: Brier is $(p-y)^2$ and log-loss is the
soft-label Bernoulli cross-entropy
$-y\log p - (1-y)\log(1-p)$ with $p$ the scalar expected-score
prediction and $y\in\{1.0,\,0.5,\,0.0\}$ -- not the 3-class Brier on
which the model was trained and not the binary log-loss that would
require dropping draws.

\subsection{Per-Rating-Band Specialization}

\noindent For each of the 14 Elo bands $[0,1000), [1000,1100), [1100,1200),
\ldots, [2100,2200), [2200,3500]$, we train an independent move model and an
independent clock model. Winner models are trained per band keyed on the
average rating of the two players. At serve time the backend routes a
request to the band matching the to-move player's
rating (move/clock) or the average of both players' ratings (winner),
which mirrors the per-record routing used in
the evaluation reported below. The trade-off with a single skill-conditioned
model like Maia-2 is clear: per-band ensembles add some maintenance cost,
but each individual model is small, specialized, and cheap to cache.

\subsection{Training}

We train each model end-to-end with Adam (fused), bf16 mixed precision,
batch size $2048$, and gradient clipping at $1.0$. The from-scratch
$1800\text{--}1900$ root move model was trained on $8{\times}$H100
GPUs; every subsequent fine-tune (the 13 move-cascade bands plus all
14 clock and 14 winner models) ran on a single NVIDIA RTX 5090.
Validation comes from a held-out month not used in training
(\S\ref{sec:data}). The 42 deployed checkpoints are produced from a
single from-scratch root model via two cascades of fine-tuning:

\begin{enumerate}[leftmargin=*,topsep=2pt,itemsep=2pt]
\item \textbf{Root model.} A single move model is trained from scratch on
  the $1800\text{--}1900$ band with ReduceLROnPlateau until convergence.
\item \textbf{Move cascade.} The remaining 13 move models are fine-tuned
  from the root move checkpoint with OneCycleLR for $1\text{--}2$ epochs
  on the target band's training shard. Transformer block weights and side
  embeddings are transferred unchanged; only the data distribution shifts.
\item \textbf{Clock and winner cascades.} All 14 clock models and all 14
  winner models are fine-tuned from the root move checkpoint with
  OneCycleLR for $1\text{--}5$ epochs. The transformer block weights are
  transferred unchanged, the output projection is reinitialized, and the
  side-feature embedding is widened to match the new arity: for the clock
  model the move model's 1D clock embedding is broadcast across the three
  clock dimensions (with the increment column scaled by $0.1$); for the
  winner model the same 1D rating embedding seeds the white/black rating
  columns, with the two clock columns scaled by $0.5$ and the increment
  column by $0.1$. After this initialization step the entire model is
  fine-tuned end-to-end.
\end{enumerate}

Concretely, only the root model is trained from scratch; every other
deployed checkpoint is reached by a short OneCycleLR fine-tune from it.

\subsection{Hyperparameters}

Table~\ref{tab:hparams} lists the shared training configuration; the
architecture is identical across the three task models (move, clock,
winner) up to the side-feature embedding arity and output projection
described above. Per-model parameter counts, weight sizes, and serving
footprint are deferred to \S\ref{sec:deployment}.

\begin{table}[h]
\centering\small
\begin{tabular}{l l}
\toprule
Hyperparameter & Value \\
\midrule
Layers & 8 \\
Embedding dim & 256 \\
Attention heads (per layer) & 8 \\
MLP width (SwiGLU) & $4 \times 256 = 1024$ \\
Dropout & 0 \\
Batch size & 2048 \\
Optimizer & Adam (fused), default $(\beta_1,\beta_2) = (0.9, 0.999)$ \\
Weight decay & 0 \\
Gradient clip & 1.0 \\
Mixed precision & bf16 \\
Root learning rate (1800-1900 from scratch) & $4\times 10^{-4}$ \\
Fine-tune base LR (OneCycleLR) & $1\times 10^{-4}$ \\
Fine-tune max LR (OneCycleLR) & $1\times 10^{-3}$ (peak), $25\times$ warmup div, $30\%$ warmup \\
Fine-tune epochs & 1-2 (move cascade); 2-5 (clock/winner cascades) \\
Root-model hardware & $8{\times}$H100 GPUs \\
Fine-tuning hardware & single NVIDIA RTX 5090 GPU \\
\bottomrule
\end{tabular}
\caption{Training hyperparameters shared across all three task models and
  all 14 bands. The 1800-1900 move model is trained from scratch with
  ReduceLROnPlateau on epoch-averaged training loss; the other 41 deployed
  checkpoints are reached by OneCycleLR fine-tuning from the root.}
\label{tab:hparams}
\end{table}

\subsection{Discussion}

\paragraph{Why per-band ensembles?}
Specialized per-band models are the less common design choice in
recent work -- Maia-2~\cite{tang2024maia2} explicitly argues for a unified
skill-conditioned model -- but they bring several pragmatic wins. Each band
fits comfortably in memory ($\sim$110\,MB checkpoint, served on CPU); the
band routing logic is trivial; and improving a single band (say, by
retraining on more recent data, or by tuning a hyperparameter for fast
games) does not risk regressions in any other. The benchmark below
suggests that for blitz, where data per band is plentiful, specialization
trades favorably against parameter sharing.

\paragraph{Joint training vs.\ a genuinely shared encoder.}
We considered a single multi-head model -- one encoder body, three task
output projections, all three losses summed during training -- which is the
design \textsc{ChessMimic} is sometimes assumed to have from a casual
reading of the architecture diagram. We did not adopt it. Per-task training
was easier to debug and to roll forward incrementally, and the fine-tuning
recipe described above captures part of the benefit: the
clock and winner models can be initialized from a move-model checkpoint,
so their transformer block weights start from the move model's solution
before being fine-tuned end-to-end. After fine-tuning those weights have
diverged, so this is shared \emph{initialization}, not shared inference-time
weights. For a larger-scale follow-up we would revisit a genuinely
shared-backbone multi-head design, with task-specific loss weights and
gradient balancing, which would reduce the deployed-model footprint by
roughly $3\times$.

\section{Experiments}
\label{sec:experiments}

\subsection{Setup}

We evaluate against Maia-2's official \texttt{blitz} checkpoint
\cite{tang2024maia2}, which is the most direct comparable model for this
setting: a unified human-imitating chess model trained on Lichess blitz,
conditioned on both players' Elo, and publicly available with
inference-ready code. We deliberately do not benchmark against engines
like Stockfish or AlphaZero, whose objective is to win rather than to
imitate, and whose top-1 move-match rates against human play are well
below human-imitation models~\cite{mcilroyyoung2020maia}. The held-out
test slice is the $n{=}4{,}775{,}033$ April 2026 sample described in
\S\ref{sec:data}.

\paragraph{Inference Details.} 
We route each record to the \textsc{ChessMimic} band matching its
$\text{elo}_\text{self}$ (move model) or the average rating (winner
model and clock model). The winner model's
3-class output is converted to scalar expected score as described in
\S\ref{sec:method}.

\paragraph{What the benchmark measures.} Every reported \textsc{ChessMimic}
number is the output of the neural model alone: the deployed common moves
database (\S\ref{sec:deployment}) is disabled for the benchmark so the
comparison to Maia-2 -- which has no such lookup -- is apples-to-apples and
isolates what the neural model has learned. The database's effect on
held-out accuracy is measured separately in \S\ref{sec:movepred}.

\subsection{Move Prediction}
\label{sec:movepred}

Table~\ref{tab:moves} summarizes top-$k$ accuracy. \textsc{ChessMimic}
gains $+3.62$ percentage points top-1 over Maia-2 averaged across the
$4{,}775{,}033$ positions, $+2.54$ top-3, and $+1.63$ top-5. The smaller gap at
top-$k$ for larger $k$ is consistent with both models concentrating most
of their probability mass on roughly the same shortlist of moves but
disagreeing more about the ordering within it.

\begin{table}[h]
\centering\small
\begin{tabular}{l c c c}
\toprule
Metric & Maia-2 [95\% CI] & \textsc{ChessMimic} [95\% CI] & $\Delta$ [95\% CI] \\
\midrule
top-1 & 0.5260 [0.5256, 0.5265] & \textbf{0.5623 [0.5618, 0.5627]} & $+0.0362\;[+0.0359, +0.0366]$ \\
top-3 & 0.8057 [0.8053, 0.8061] & \textbf{0.8310 [0.8307, 0.8314]} & $+0.0254\;[+0.0251, +0.0256]$ \\
top-5 & 0.8969 [0.8966, 0.8972] & \textbf{0.9132 [0.9129, 0.9134]} & $+0.0163\;[+0.0161, +0.0165]$ \\
\bottomrule
\end{tabular}
\caption{Move prediction on the month-wide Lichess April 2026 blitz
  sample (one randomly chosen non-opening ply per game,
  $n=4{,}775{,}033$), with closed-form $95\%$ intervals: Wilson for the
  per-model proportions, normal-approximation paired SE for $\Delta$. All
  three $\Delta$ intervals are entirely above zero. The one-ply-per-game
  design removes within-game correlation, so the i.i.d.\ analytic
  interval is well-justified; residual between-game clustering
  (repeated players, openings) is a smaller residual caveat.}
\label{tab:moves}
\end{table}

\paragraph{Common moves database has negligible impact on top-1.}
As part of model training, we calculate an empirical move distribution
database for common positions. Like the models, these empirical distributions
are calculated separately per-band. The common positions primarily occur during
openings and endgames.
In the 1e4.ai demo, we check if the position is in the empirical database,
and if it is then we serve a move from its distribution. Otherwise,
we sample from the move model's distribution to serve the move.
To investigate the impact of serving from the empirical database, we re-scored the
$4{,}775{,}033$-position held-out slice using 1e4.ai's method of first
consulting the database and then using the move model.
Only $\mathbf{3.2\%}$ of records overall are in the database
(peaking at $7.5\%$ in the 1700-1800 band), and on those records the
empirical distribution's top-1 ($52.4\%$) is slightly lower than the move model's top-1
($53.0\%$). Enabling the database only changes the overall top-1 by
$-0.02$\,pp ($0.5623 \to 0.5621$) on this slice, well within noise.

The \textsc{ChessMimic}--Maia-2 gap widens toward the top of the rating
range (Table~\ref{tab:moves_per_band}), from roughly $+3$\,pp through
the lower and middle bands to $+6.1$\,pp at 2200+. Most of this widening
is a property of Maia-2 rather than \textsc{ChessMimic}: Maia-2's
per-band top-1 plateaus near 1900--2000 and slips by 2200+, whereas
\textsc{ChessMimic} improves monotonically across all 14 bands. We do
not train a controlled unified-backbone \textsc{ChessMimic} variant
(\S\ref{sec:limitations}), so we cannot attribute the gap to any single
factor among architecture, data, loss, and preprocessing.

\begin{figure}[h]
\centering
\includegraphics[width=\textwidth]{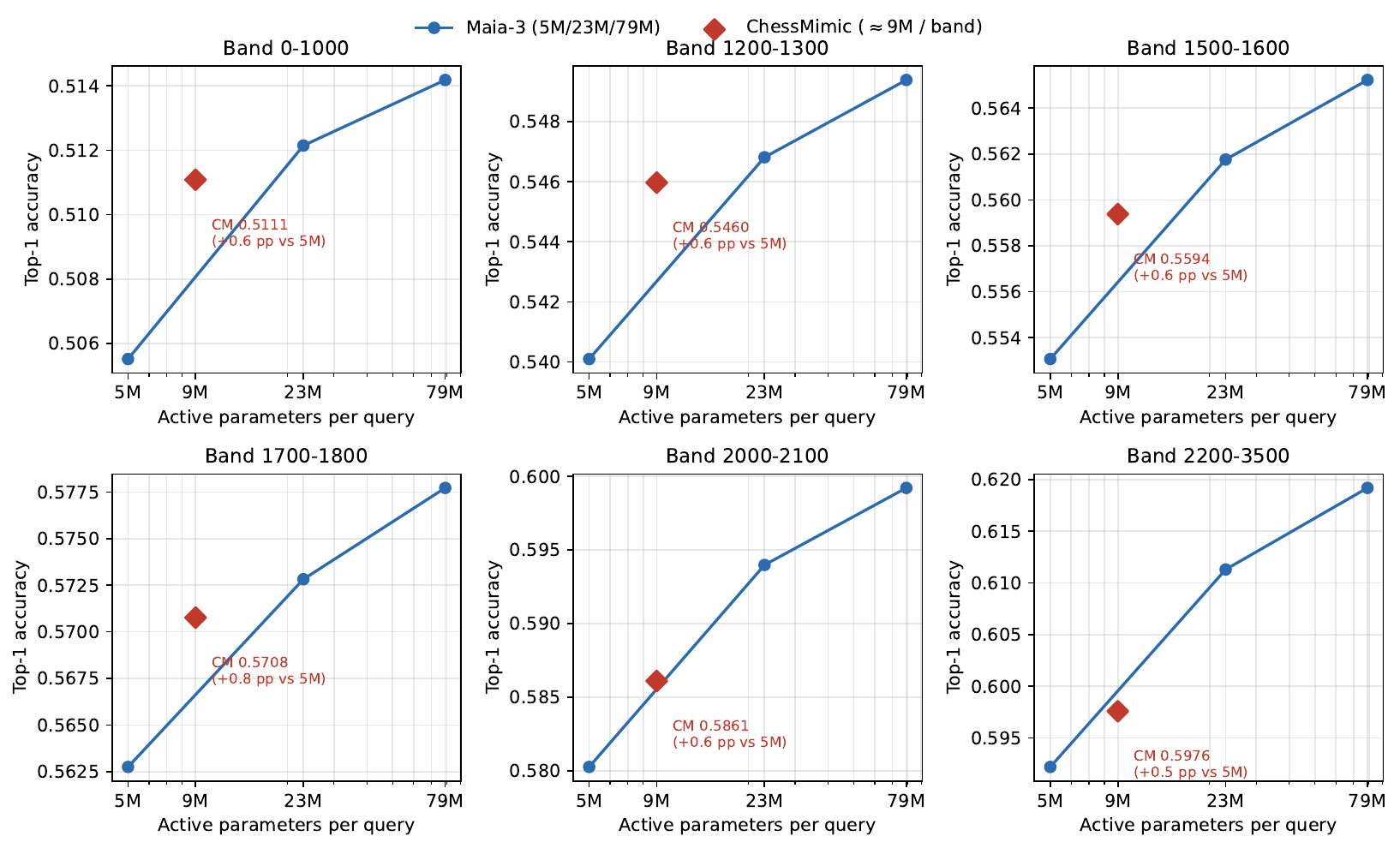}
\caption{Per-band top-1 move accuracy vs active parameters per query for
  Maia-3 (5M / 23M / 79M, blue line) and \textsc{ChessMimic}
  ($\approx$9M per band, red diamond), drawn straight from
  Table~\ref{tab:moves_per_band} on the bot-filtered April 2026
  held-out slice ($n=4{,}775{,}033$). The $x$-axis is log-spaced.
  Six representative bands span the rating range; the same shape
  appears in all 14. \textsc{ChessMimic}'s 9M per-band model lies
  \emph{above} the Maia-3 scaling line at 9M in the bottom four panels
  (i.e., its accuracy beats what a parameter-matched Maia-3 would
  achieve under log-linear interpolation between 5M and 23M), is
  essentially on the line at 2000-2100, and falls slightly below it at
  2200+. This is the parameter-efficiency picture behind the per-band
  numbers in Table~\ref{tab:moves_per_band}: per-band specialisation
  buys a few hundredths in the low-and-mid bands at a fraction of the
  per-query parameter cost, but Maia-3's unified backbone keeps
  scaling faster at the rating extreme.}
\label{fig:param_vs_accuracy}
\end{figure}

The Maia-3 columns place \textsc{ChessMimic} in parameter-scaling
context. At $\approx$9M active parameters per query it lands between
Maia-3-5M and Maia-3-23M in every one of the 14 bands -- above the 5M
model and only modestly below the 23M, which carries
$\approx$2.5$\times$ the active parameters -- roughly where a 9M-active
model should fall on the scaling curve. The largest unified Chessformer,
Maia-3-79M, tops every row, and its margin over \textsc{ChessMimic}
grows with rating, from $+0.3$\,pp at 0-1000 to $+2.2$\,pp at 2200+.
Figure~\ref{fig:param_vs_accuracy} plots this parameter-vs-accuracy
trade-off for six representative bands.

\begin{table}[h]
\centering\small
\setlength{\tabcolsep}{4pt}
\begin{tabular}{l r r r r r r}
\toprule
Band & $n$ & Maia-2 & Maia-3-5M & Maia-3-23M & Maia-3-79M & \textsc{ChessMimic} \\
\midrule
0-1000      & 275{,}528 & 0.4782 & 0.5055 & 0.5121 & \textbf{0.5142} & 0.5111 \\
1000-1100   & 195{,}496 & 0.4986 & 0.5273 & 0.5334 & \textbf{0.5357} & 0.5316 \\
1100-1200   & 245{,}714 & 0.5091 & 0.5332 & 0.5407 & \textbf{0.5437} & 0.5397 \\
1200-1300   & 307{,}370 & 0.5141 & 0.5401 & 0.5468 & \textbf{0.5494} & 0.5460 \\
1300-1400   & 370{,}668 & 0.5175 & 0.5426 & 0.5496 & \textbf{0.5535} & 0.5492 \\
1400-1500   & 421{,}170 & 0.5230 & 0.5476 & 0.5561 & \textbf{0.5596} & 0.5533 \\
1500-1600   & 475{,}317 & 0.5273 & 0.5531 & 0.5618 & \textbf{0.5652} & 0.5594 \\
1600-1700   & 490{,}137 & 0.5330 & 0.5601 & 0.5694 & \textbf{0.5726} & 0.5673 \\
1700-1800   & 489{,}493 & 0.5350 & 0.5627 & 0.5728 & \textbf{0.5777} & 0.5708 \\
1800-1900   & 451{,}959 & 0.5392 & 0.5696 & 0.5807 & \textbf{0.5854} & 0.5771 \\
1900-2000   & 374{,}946 & 0.5438 & 0.5764 & 0.5883 & \textbf{0.5929} & 0.5833 \\
2000-2100   & 279{,}212 & 0.5421 & 0.5803 & 0.5940 & \textbf{0.5992} & 0.5861 \\
2100-2200   & 179{,}370 & 0.5429 & 0.5851 & 0.6011 & \textbf{0.6062} & 0.5908 \\
2200-3500   & 218{,}653 & 0.5370 & 0.5922 & 0.6113 & \textbf{0.6192} & 0.5976 \\
\bottomrule
\end{tabular}
\caption{Per-band top-1 move accuracy on the month-wide April 2026
  bot-filtered blitz sample (one ply per game; $n$ is the natural
  per-band count). \textsc{ChessMimic} outperforms Maia-2 in all 14
  bands (gap from $+3.0$\,pp at 1400-1500 to $+6.1$\,pp at 2200+,
  monotone above 1500), and every per-band Maia-2 vs \textsc{ChessMimic}
  Wilson 95\% interval lies strictly above zero. The Maia-3 columns
  show where \textsc{ChessMimic}'s per-band specialisation lands against
  the strongest released open human-modelling baselines at each rating.}
\label{tab:moves_per_band}
\end{table}

\subsection{Head-to-head on the ALLIE 2022 blitz test slice}
\label{sec:chessformer_compare}

The Chessformer / Maia-3 paper~\cite{monroe2026chessformer} reports
move-matching accuracy on the $884{,}049$-position ALLIE 2022 blitz test
set of Zhang~et~al.~\cite{zhang2025allie}: $\sim$18k held-out Lichess
2022 blitz games (downsampled to roughly equal games per 100-Elo bin),
with opening moves dropped and every position from the first time the
mover's clock falls below 30\,s excluded. Because \textsc{ChessMimic}
trains only on 2024-09 through 2025-08 (\S\ref{sec:data}), the 2022
slice is clean held-out data for it. Maia-3's checkpoints are released
at \texttt{github.com/CSSLab/maia3}, so we run all three sizes
(5M / 23M / 79M) directly on the reconstructed slice; among the models
we re-run, this is a same-positions comparison (the published
\textsc{Allie}-Policy number is on the full set and is included only
for context, not for direct apples-to-apples reading). We mirror
Zhang~et~al.'s slice construction verbatim and do \emph{not} apply our
bot-account filter (\S\ref{sec:data}) here, since doing so would
diverge from the published reference and invalidate the parity check
against Chessformer's reported Maia-3 numbers; the held-out April
2026 slice in \S\ref{sec:movepred} uses the bot-filtered construction.

The frozen position list ships only in the Maia-3 training repository,
so we \emph{reconstruct} the slice from the public ALLIE test games
(HuggingFace \texttt{yimingzhang/allie-data}) by applying those two
filters, and validate the reconstruction by cross-checking model
numbers against the Chessformer paper. Our reconstruction is
$849{,}730$ positions ($96\%$ of the reported $884{,}049$; the residual
is game-set and clock-reconstruction edge differences). On it our
Maia-2 scores $0.5219$ top-1 -- within $0.2$\,pp of the $0.520$ that
Chessformer reports for Maia-2 on the official set; our
directly-evaluated Maia-3-5M / 23M / 79M land at
$0.5565$ / $0.5677$ / $0.5729$, all within $0.25$\,pp of the reported
$0.554$ / $0.566$ / $0.571$, and in the same direction (our $96\%$
reconstruction is marginally easier across all four anchor models).
That parity is the basis for the head-to-head comparison: same
reconstructed game pool and filters for the models we directly
evaluate, same metric, and four anchor models agree to two decimal
places. \textsc{ChessMimic}'s 14 per-band move models span Elo
$0$ to $3500$, so every position in the slice is scored with no
Elo-coverage gap.

\begin{table}[h]
\centering\small
\begin{tabular}{lrrrrl}
\toprule
Model & Params & Top-1 & Top-3 & Top-5 & Positions \\
\midrule
\multicolumn{6}{l}{\emph{Reported on the full set (different slice; top-1 only)}} \\
\textsc{Allie}-Policy~\cite{zhang2025allie} & --   & 0.557 & --    & --    & 884{,}049 (full) \\
\midrule
\multicolumn{6}{l}{\emph{Directly evaluated by this work on the $849{,}730$-position reconstruction}} \\
Maia-2                                    & 23M  & 0.522 & 0.800 & 0.893 & 849{,}730 \\
Maia-3-5M                                 &  5M  & 0.557 & 0.827 & 0.910 & 849{,}730 \\
Maia-3-23M                                & 23M  & 0.568 & 0.836 & 0.917 & 849{,}730 \\
Maia-3-79M                                & 79M  & 0.573 & 0.839 & 0.919 & 849{,}730 \\
\textsc{ChessMimic}                       & $\approx$9M / band & \textbf{0.562} & \textbf{0.830} & \textbf{0.911} & 849{,}730 \\
\bottomrule
\end{tabular}
\caption{Move top-1 / top-3 / top-5 on the ALLIE 2022 blitz test
  slice. Upper block: \textsc{Allie}-Policy as reported
  by~\cite{zhang2025allie} on the full $884{,}049$-position set, top-1
  only, included for context but \emph{not} directly comparable to the
  reconstruction rows below. Lower block: this work, run on our
  $849{,}730$-position reconstruction of the slice ($96\%$ of the
  reported $884{,}049$; the residual is game-set and
  clock-reconstruction edge differences). The reconstruction is
  validated by Maia-2 reaching $0.522$ on it vs the $0.520$ Chessformer
  reports on the full set, and by all three Maia-3 sizes reproducing
  the reported $0.554$ / $0.566$ / $0.571$ to within $0.25$\,pp.
  \textsc{ChessMimic} is scored on every position (all 14 Elo bands).}
\label{tab:chessformer}
\end{table}

\textsc{ChessMimic} reaches $\mathbf{0.562}$ top-1 over the full
reconstructed slice, against a $0.522$ Maia-2 anchor -- a $+4.0$\,pp
\textsc{ChessMimic} lead, the same direction and magnitude as on the
independent April 2026 month-wide set (\S\ref{sec:movepred}), and
\textsc{ChessMimic} outperforms Maia-2 in every one of the 14 Elo bands
(per-band $\Delta$ from $+3.0$\,pp at 1100-1200 to $+6.2$\,pp at 2200+;
even the hard sub-1000 band, where moves are noisiest, goes $0.502$ vs
$0.466$). At $\approx$9M active parameters per query, \textsc{ChessMimic}'s
$0.562$ exceeds the directly-evaluated Maia-3-5M ($0.557$) and the
reported \textsc{Allie}-Policy ($0.557$), just below Maia-3-23M
($0.568$), and below Maia-3-79M's $0.573$ -- the largest unified model
at $\approx$8x the per-band parameter count. The per-query comparison
is apples-to-apples; the deployed move ensemble carries 14
per-band checkpoints for a total of $\approx 132$M parameters
(Table~\ref{tab:footprint}), so the deployment footprint is closer to
Maia-3-23M than to Maia-3-5M. The same shape carries to
top-3 and top-5: \textsc{ChessMimic}'s $0.830$ / $0.911$ sit just above
Maia-3-5M's $0.827$ / $0.910$, just below Maia-3-23M's $0.836$ /
$0.917$, and the gap to Maia-3-79M narrows from $-1.1$\,pp at top-1 to
$-0.8$\,pp at top-5. Chessformer does not report value-head calibration
or AUC, so the expected-score comparison (\S\ref{sec:winpred}) still
cannot be carried across to that family.

\subsection{Outcome / expected-score prediction}
\label{sec:winpred}

Table~\ref{tab:winners} reports outcome prediction. Maia-2's published
value head is trained on game outcomes~\cite{tang2024maia2}, but it is
a scalar auxiliary output without clock inputs and was not designed or
evaluated as a calibrated clock-aware human-blitz expected-score
forecaster; we include it because it is the only released
human-modeling baseline with a value output, but treat it as
contextual rather than as the primary baseline. The informative
comparisons for this task are the lightweight tabular baselines in
Table~\ref{tab:winner_baselines}.

\begin{table}[h]
\centering\small
\begin{tabular}{l c c}
\toprule
Metric & Maia-2 [95\% CI] & \textsc{ChessMimic} [95\% CI] \\
\midrule
Brier $\downarrow$ & 0.2787 [0.2785, 0.2789] & \textbf{0.1837 [0.1835, 0.1838]} \\
Log-loss $\downarrow$ & 0.8501 [0.8493, 0.8510] & \textbf{0.5647 [0.5644, 0.5651]} \\
AUC (draws excluded, $n{=}4{,}565{,}557$) $\uparrow$ & 0.5012 [0.5007, 0.5017] & \textbf{0.7768 [0.7764, 0.7772]} \\
\bottomrule
\end{tabular}
\caption{Win prediction with closed-form $95\%$ intervals (normal-approx
  SE for Brier/log-loss, DeLong for AUC). Target is the side-to-move's
  expected score ($1.0 / 0.5 / 0.0$ for win/draw/loss). The Maia-2 vs
  \textsc{ChessMimic} difference on Brier is $-0.0950$ [$-0.0953$, $-0.0948$]
  and on log-loss is $-0.285$ [$-0.286$, $-0.284$], both entirely below
  zero. At this sample size Maia-2's AUC is pinned at $\approx\!0.50$
  (just barely separable from chance and far below \textsc{ChessMimic}'s
  $0.78$): its scalar value output is trained on game
  outcomes~\cite{tang2024maia2} but receives no clock conditioning and
  was not evaluated as a clock-aware blitz expected-score forecaster,
  which is the axis this task asks about.}
\label{tab:winners}
\end{table}

The decile calibration in Table~\ref{tab:calibration} shows the
qualitative shape of the gap. \textsc{ChessMimic}'s calibration is
monotone: bin 1 predicts $0.085$ and observes $0.080$; bin 10 predicts
$0.923$ and observes $0.932$. Maia-2's decile calibration is flat
across the same axis: bin 1 predicts $0.129$ but observes $0.543$;
bin 10 predicts $0.882$ but observes only $0.548$. The Maia-2
predictions span almost the entire $[0, 1]$ interval (decile mean range
$0.13$ to $0.88$), so the issue is not saturation: it is that the
predicted ranking on this task -- clock-aware human-blitz outcomes --
is essentially flat, consistent with Maia-2's value head having been
trained on a different objective.

\begin{table}[h]
\centering\small
\begin{tabular}{r r r r r r r}
\toprule
& \multicolumn{3}{c}{Maia-2} & \multicolumn{3}{c}{\textsc{ChessMimic}} \\
\cmidrule(lr){2-4}\cmidrule(lr){5-7}
Decile & $n$ & predicted & empirical & $n$ & predicted & empirical \\
\midrule
1  & 477{,}504 & 0.129 & 0.543 & 477{,}504 & 0.085 & 0.080 \\
3  & 477{,}504 & 0.414 & 0.492 & 477{,}504 & 0.374 & 0.367 \\
5  & 477{,}503 & 0.499 & 0.498 & 477{,}503 & 0.481 & 0.481 \\
7  & 477{,}503 & 0.563 & 0.496 & 477{,}503 & 0.564 & 0.571 \\
10 & 477{,}503 & 0.882 & 0.548 & 477{,}503 & 0.923 & 0.932 \\
\bottomrule
\end{tabular}
\caption{Decile calibration of expected-score predictions,
  $n{=}4{,}775{,}033$ ($\approx$477{,}503 per decile). Selected rows
  shown for compactness; the full table is monotone for
  \textsc{ChessMimic} and approximately flat for Maia-2.}
\label{tab:calibration}
\end{table}

\paragraph{Per-band Brier.}
Table~\ref{tab:winners_per_band} reports outcome Brier per side-to-move
band; \textsc{ChessMimic} beats Maia-2 in every band by a wide margin
($\Delta \in [-0.106, -0.092]$, every per-band interval strictly below
zero). All 14 winner-model bands have a trained
checkpoint; winner inference is routed by the average of the two players'
ratings, matching production.

\begin{table}[h]
\centering\small
\begin{tabular}{l r r r r}
\toprule
Side-to-move band & $n$ & Maia-2 Brier $\downarrow$ & \textsc{ChessMimic} Brier $\downarrow$ & $\Delta$ \\
\midrule
0-1000      & 275{,}528 & 0.283 & \textbf{0.178} & $-0.106$ \\
1000-1100   & 195{,}496 & 0.280 & \textbf{0.183} & $-0.097$ \\
1100-1200   & 245{,}714 & 0.282 & \textbf{0.184} & $-0.097$ \\
1200-1300   & 307{,}370 & 0.281 & \textbf{0.185} & $-0.097$ \\
1300-1400   & 370{,}668 & 0.281 & \textbf{0.185} & $-0.095$ \\
1400-1500   & 421{,}170 & 0.281 & \textbf{0.188} & $-0.093$ \\
1500-1600   & 475{,}317 & 0.279 & \textbf{0.187} & $-0.093$ \\
1600-1700   & 490{,}137 & 0.279 & \textbf{0.185} & $-0.094$ \\
1700-1800   & 489{,}493 & 0.278 & \textbf{0.184} & $-0.094$ \\
1800-1900   & 451{,}959 & 0.278 & \textbf{0.184} & $-0.094$ \\
1900-2000   & 374{,}946 & 0.277 & \textbf{0.183} & $-0.095$ \\
2000-2100   & 279{,}212 & 0.277 & \textbf{0.182} & $-0.096$ \\
2100-2200   & 179{,}370 & 0.273 & \textbf{0.180} & $-0.093$ \\
2200-3500   & 218{,}653 & 0.266 & \textbf{0.174} & $-0.092$ \\
\bottomrule
\end{tabular}
\caption{Per-band outcome Brier (lower is better) on the
  month-wide April 2026 sample ($n$ is the natural per-band count).
  Rows are by side-to-move band;
  winner inference is routed by the two players' average rating, so a row
  may be served by a neighboring band's winner model when the opponent is
  far from the side-to-move's band. All 14 winner bands have a trained
  checkpoint.}
\label{tab:winners_per_band}
\end{table}

\paragraph{What drives the gap?}
Two factors are consistent with what we observe. First,
\textsc{ChessMimic}'s winner model uses a 3-class softmax over W/D/L
with explicit draw mass; Maia-2's value head is a scalar auxiliary
output trained on game outcomes~\cite{tang2024maia2} but without
clock inputs, and on our held-out clock-aware blitz slice its
\texttt{win\_prob} output is essentially flat against eventual results
(AUC $\approx\!0.50$). As an illustration, consider a forced back-rank
mate position (Lichess game \texttt{s6wp7q6M}, after 25.Qd8+): the
only legal Black move is a queen recapture which is itself immediately
followed by \texttt{Rxd8\#}. Maia-2 outputs Black expected score
$= 1.0000$ on this position -- consistent with reading material
balance after the trade -- while \textsc{ChessMimic} outputs $0.0084$,
correctly registering the mating pattern that resolved in the training
data. Second, \textsc{ChessMimic}'s winner model receives both players'
clock states and the increment, which carry real signal about who is
likely to flag (out-of-time loss) in blitz. Maia-2 was designed without
clock inputs, which is a structural choice rather than a benchmarking
unfairness; we mention it because it explains a portion of the gap.

For move prediction the gap is much smaller (single-digit percentage
points) and the qualitative behavior is similar between the two models --
both correctly assign almost all of their probability to the same small
set of plausible human moves in most positions. The winner-model gap is
structurally different: Maia-2's $\text{win\_prob}$ is essentially
uncorrelated with outcomes on this sample (AUC $0.50$, the chance
baseline) where \textsc{ChessMimic}'s tracks them (AUC $0.78$). The
next paragraph shows that recalibrating Maia-2 cannot close it -- this
is a discrimination gap, not a calibration one.

\paragraph{Is Maia-2 just a weak baseline?}
To check whether the winner model adds signal beyond obvious tabular
features -- and to test whether Maia-2's value head is merely
miscalibrated rather than uninformative -- we score five simple
baselines on the same month-wide slice
(Table~\ref{tab:winner_baselines}). The baselines are fit with 5-fold
cross-fit predictions so no position is scored by a model
trained on it. The two binary-logistic baselines fit on decisive games
(win$=$1/loss$=$0) and predict $P(\text{win})$ as the target.
The draw-aware multinomial baseline fits a $3$-class W/D/L logit
with features rating gap, clocks, increment, signed
material, ply count, legal-move count, side-to-move colour, and
per-side piece counts (pawns, knights, bishops, rooks, queens), then
calculates the expected score $P(\text{win}) + 0.5\,P(\text{draw})$. 
The isotonic baseline
recalibrates Maia-2's raw \texttt{win\_prob} against the
$1.0/0.5/0.0$ target. We note four observations: (i) the closed-form Elo
expectation is weak (Brier $0.237$, AUC $0.56$); adding clock and
material each improves both scores (Brier $0.208$, AUC $0.70$), and
folding draws into a 3-class multinomial gives a small additional
gain (Brier $0.207$, AUC $0.71$).
(ii) \textsc{ChessMimic} still beats the strongest of the simple baselines by
$0.023$ Brier and $0.07$ AUC, so the transformer captures
board-and-clock structure beyond rating gap, material, clocks, and
explicit piece counts. (iii) Recalibrating Maia-2 improves its Brier
($0.279\to0.237$) but leaves AUC essentially at chance
($0.501\to0.513$). Maia-2's value output has a discrimination
problem on this task, not just a calibration one.
(iv) \textsc{ChessMimic} wins every row on all three metrics by a statistically significant amount.

\begin{table}[h]
\centering\small
\begin{tabular}{l c c c}
\toprule
Predictor & Brier $\downarrow$ & Log-loss $\downarrow$ & AUC $\uparrow$ \\
\midrule
Rating-only Elo expectation        & 0.2374 [0.2373, 0.2375] & 0.6905 [0.6904, 0.6907] & 0.5550 [0.5544, 0.5555] \\
Rating $+$ clock logistic          & 0.2281 [0.2280, 0.2282] & 0.6707 [0.6705, 0.6709] & 0.6192 [0.6187, 0.6197] \\
Material $+$ rating $+$ clock logistic & 0.2084 [0.2083, 0.2086] & 0.6265 [0.6262, 0.6268] & 0.7014 [0.7009, 0.7019] \\
Draw-aware multinomial (W/D/L)     & 0.2070 [0.2069, 0.2071] & 0.6231 [0.6228, 0.6234] & 0.7074 [0.7069, 0.7079] \\
Maia-2 (isotonic-recalibrated)     & 0.2372 [0.2372, 0.2373] & 0.6892 [0.6891, 0.6893] & 0.5127 [0.5122, 0.5132] \\
Maia-2 (raw)                       & 0.2787 [0.2785, 0.2789] & 0.8501 [0.8493, 0.8510] & 0.5012 [0.5007, 0.5017] \\
\textsc{ChessMimic} winner         & \textbf{0.1837 [0.1835, 0.1838]} & \textbf{0.5647 [0.5644, 0.5650]} & \textbf{0.7768 [0.7764, 0.7772]} \\
\bottomrule
\end{tabular}
\caption{Outcome-prediction baselines on the same month-wide April 2026
  slice, with closed-form $95\%$ intervals (normal-approx SE for
  Brier/log-loss, DeLong for AUC). Fitted baselines use 5-fold cross-fit
  predictions (no row scored by a model trained on it). AUC excludes
  draws ($n{=}4{,}565{,}557$). \textsc{ChessMimic} beats the strongest
  non-neural baseline (the draw-aware W/D/L multinomial) by $0.023$
  Brier and $0.07$ AUC; recalibrating Maia-2 barely moves its AUC,
  showing the value-head gap is discrimination, not calibration.}
\label{tab:winner_baselines}
\end{table}

\subsection{Self-Play Rating Sanity Check}

As a sanity check that the rating conditioning actually controls strength,
we ran 1{,}000-game self-play matches between models in different
rating bands using a Glicko-2~\cite{glickman2012glicko2} updater initialized
at 1500 / RD 500. Two 1200-band models converge to a stable rating
difference within their RD; two 1800-band models likewise; and a
1200-vs-2000 match produces a $\sim$650-point separation in the expected
direction. We use this elsewhere in development to spot when a band model
degrades after a retrain. The Glicko-2 numbers themselves are not
comparable to Lichess ratings: Glicko/Elo carry meaning only within a
player pool, and our self-play pool contains just the two models
being compared (anchored to the arbitrary $1500$-point starting
rating), while Lichess ratings come from a much larger open pool.
Only the relative gap (e.g., the $\sim\!650$ points between the
$1200$ and $2000$ bands) is the comparable quantity; the rating
values themselves are not.

\subsection{Stress Tests on Clock and Time Pressure}
\label{sec:clock_studies}

The benchmark above shows that \textsc{ChessMimic}'s aggregate Brier on
outcome prediction is much better than Maia-2's. This subsection asks more
specific questions about the time-pressure dimension of that result. All
analyses below use the $100{,}000$-position per-band-stratified
sub-sample of the April 2026 held-out set (\S\ref{sec:data};
$7{,}143$ positions per band), except where noted.

\begin{figure}[h]
\centering
\includegraphics[width=0.85\textwidth]{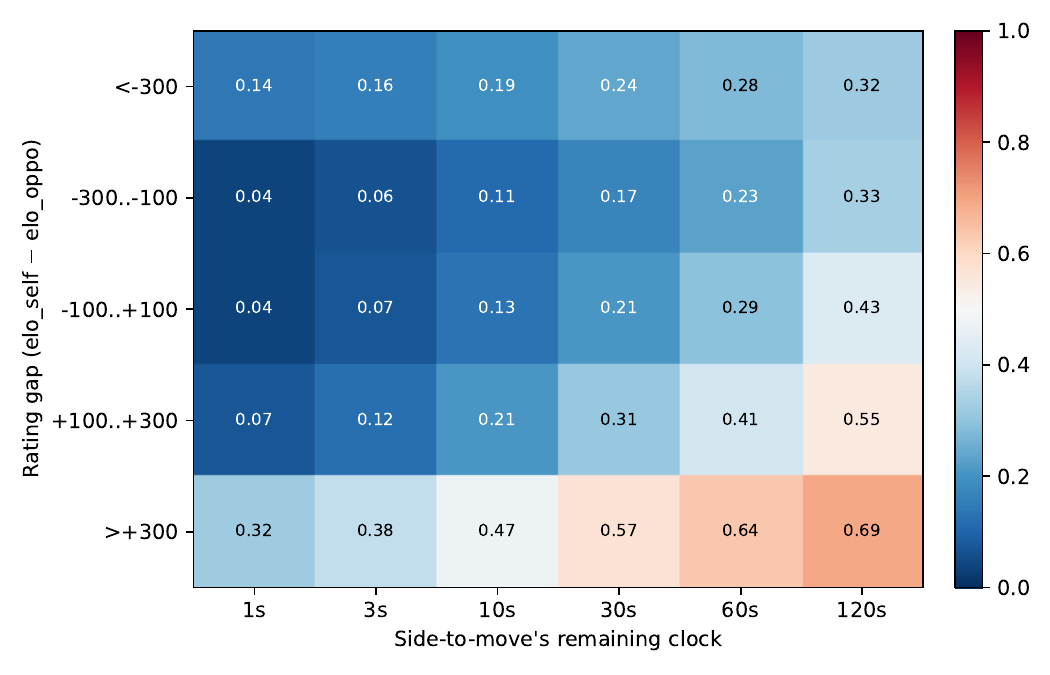}
\caption{Mean predicted counterfactual expected score for the
  side-to-move when its clock is overridden, opponent's clock held at
  the original recorded value. Each cell is a mean over records in
  that (rating gap, override clock) cell. The central
  $-100\ldots+100$ bucket has $n{=}90{,}283$ and is comfortably
  estimated; the $\Delta > +300$ ($n{=}220$) and $< -300$ ($n{=}225$)
  buckets are sparse because rating gaps that wide are rare in real
  games, so the extremes should be read more cautiously.}
\label{fig:cf_winprob}
\end{figure}

\paragraph{(A) Clock-conditioned expected score.}
For each held-out position we perform seven counterfactual queries on the
winner model: we keep the opponent's clock at its original value, but set
the side-to-move's clock at
$\{1, 3, 10, 30, 60, 120, \text{original}\}$ seconds. We then stratify by the
rating gap $\Delta = \text{elo}_\text{self} - \text{elo}_\text{opp}$
(Figure~\ref{fig:cf_winprob}). For each rating gap, the expected score
decreases monotonically as the remaining clock decreases.
The model's predictions reflect the realism of severe human time 
pressure eroding even a substantial rating advantage.

\begin{figure}[h]
\centering
\includegraphics[width=\textwidth]{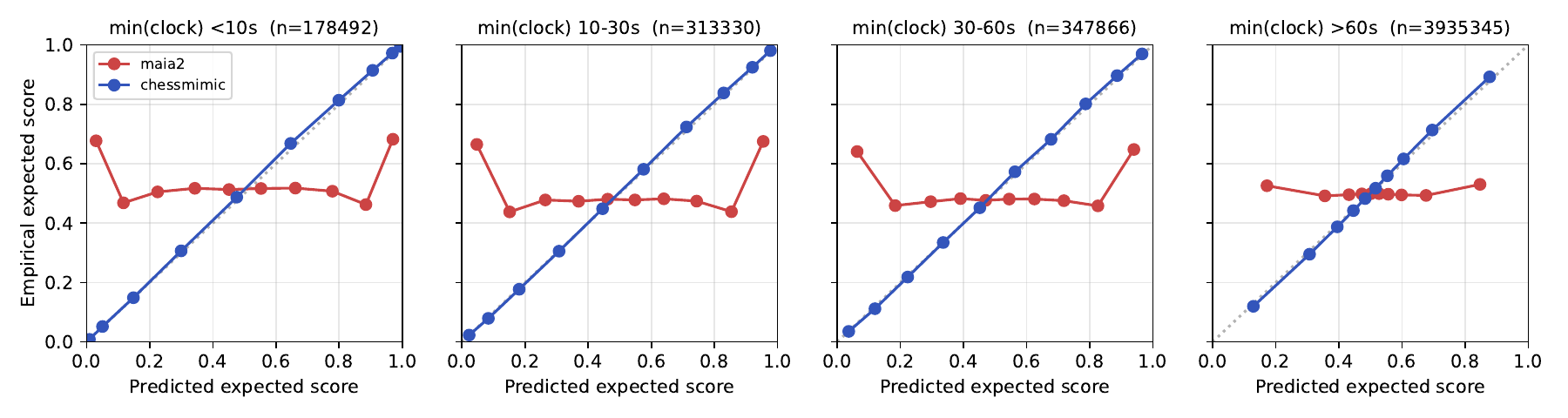}
\caption{Reliability diagram for both models, stratified by
  $\min(\text{white\_clock\_s}, \text{black\_clock\_s})$.
  \textsc{ChessMimic} (blue) tracks the diagonal in every clock-pressure
  regime; Maia-2 (red) is nearly flat in all four. The gap is largest in
  the $<\!10\,\text{s}$ bucket where time-pressure outcomes dominate.}
\label{fig:calibration_by_clock}
\end{figure}

\paragraph{(B) Winner-model calibration stratified by clock pressure.}
Bucketing the held-out positions by $\min(\text{white\_clock\_s},
\text{black\_clock\_s})$ -- the minimum of the two clocks -- separates
clock-pressured positions from comfortable ones. Within each clock-pressure
bucket, \textsc{ChessMimic}'s Brier is far below Maia-2's
(Table~\ref{tab:brier_by_clock}); the gap actually widens in the
$<\!10$-second bucket (0.09 vs 0.32), because much of the outcome at
that point is determined by who runs out of time -- a signal Maia-2 has
no input for. The reliability diagrams in
Figure~\ref{fig:calibration_by_clock} show that \textsc{ChessMimic} tracks
the diagonal in every regime, while Maia-2 is roughly flat in every regime.

\begin{table}[h]
\centering\small
\begin{tabular}{lrcc}
\toprule
$\min(\text{clocks})$ & $n$ & Maia-2 Brier $\downarrow$ & \textsc{ChessMimic} Brier $\downarrow$ \\
\midrule
$<10$\,s     & 178{,}492   & 0.324 & \textbf{0.091} \\
$10\text{--}30$\,s   & 313{,}330 & 0.309 & \textbf{0.114} \\
$30\text{--}60$\,s   & 347{,}866 & 0.303 & \textbf{0.133} \\
$>60$\,s     & 3{,}935{,}345 & 0.272 & \textbf{0.198} \\
\bottomrule
\end{tabular}
\caption{Brier score by clock-pressure bucket (lower is better),
  bucketing by the lower of the two clocks. Computed on the full
  month-wide set ($n{=}4{,}775{,}033$), since it is an $O(n)$ reduction
  over per-position predictions and needs no model re-query.
  \textsc{ChessMimic}'s gain over Maia-2 is consistent across regimes and
  \emph{strongest} in the most-pressured bucket.}
\label{tab:brier_by_clock}
\end{table}

\paragraph{(C) Per-ply thinking-time correlation vs.\ ALLIE.}
We compute the predicted thinking time per ply as
$E[t] = \sum_i P(\text{bucket}_i) \cdot \text{center}(\text{bucket}_i)$
from the clock model's bucketed distribution, and compare it to the actual
seconds the moving side spent on the move
(\texttt{pre\_clock} $-$ \texttt{post\_clock} $+$ \texttt{increment}). For
an apples-to-apples comparison we apply the three filters ALLIE reports
(\cite{zhang2025allie}, \S 4.1): drop the first 5 full-moves of every game
($=$ 10 plies, requiring $|\text{move\_history}| \geq 10$), drop any move
with the moving side under 30 seconds on the clock, and do not filter on
the thinking time itself (so zero-time premoves are retained, matching
ALLIE's MSE regression which sees them in training). This leaves
$n{=}89{,}299$ records. The overall Pearson $r$ is $0.41$
(Spearman $\rho = 0.50$; MAE $4.10$\,s) vs.\ ALLIE's reported $r{=}0.70$
on its 884k-position 2022 blitz slice.

We do not attribute this to training-set scale -- across the 12
monthly dumps 2024-09 through 2025-08 we process approximately
$4.5 \times 10^{8}$ Lichess rated blitz games in total (each monthly
dump contributes on the order of $37$--$40$\,M rated blitz games, as
in the April 2026 held-out month with $37.58$\,M), and each per-band
clock model is trained on the games (and the position-level records
derived from them) that fall in its $100$-Elo window. The exact
per-band games count depends on the dump months and the rating-band
mass, but the higher-density middle bands sit comfortably in the tens of
millions of games -- comparable in order of magnitude to ALLIE's
$91$M-game training set. Two structural differences are more likely to explain 
the gap.

\textbf{(i) Output parameterization.} \textsc{ChessMimic}'s clock head
outputs a $K{=}30$ bucketed probability distribution (trained with masked
Brier) and our predicted think time is its expectation
$E[t] = \sum_i P_i \cdot \mu_i$, where $\mu_i$ is the empirical mean of
training-time think times in bucket $i$. This is a different output
parameterization from ALLIE's pondering head, which is a \emph{scalar
regression} head trained with MSE against ground-truth think times: Brier on
a long-tailed class distribution tolerates timid under-confidence on the
rare long-think classes in a way MSE does not, since MSE blows up on missed 
large values whereas Brier caps each per-class contribution at $1$. 

\textbf{(ii) Single skill-conditioned model vs.\ per-band ensemble.}
ALLIE is a single skill-conditioned model trained on the full Elo
spectrum, so its predictions can use cross-band variance in how
different-rated players use time. \textsc{ChessMimic}'s per-band clock
models each see only $\sim$100\,Elo of variance in their training data;
the cross-band signal, which is real and large, is unavailable to any
single band's model. The per-band $r$ ranges from $0.38$ to $0.46$ with
no clear monotone trend across Elo. We report this as a measurable
per-ply think-time signal but \emph{not} as a result that matches
ALLIE.

\begin{figure}[h]
\centering
\includegraphics[width=0.62\textwidth]{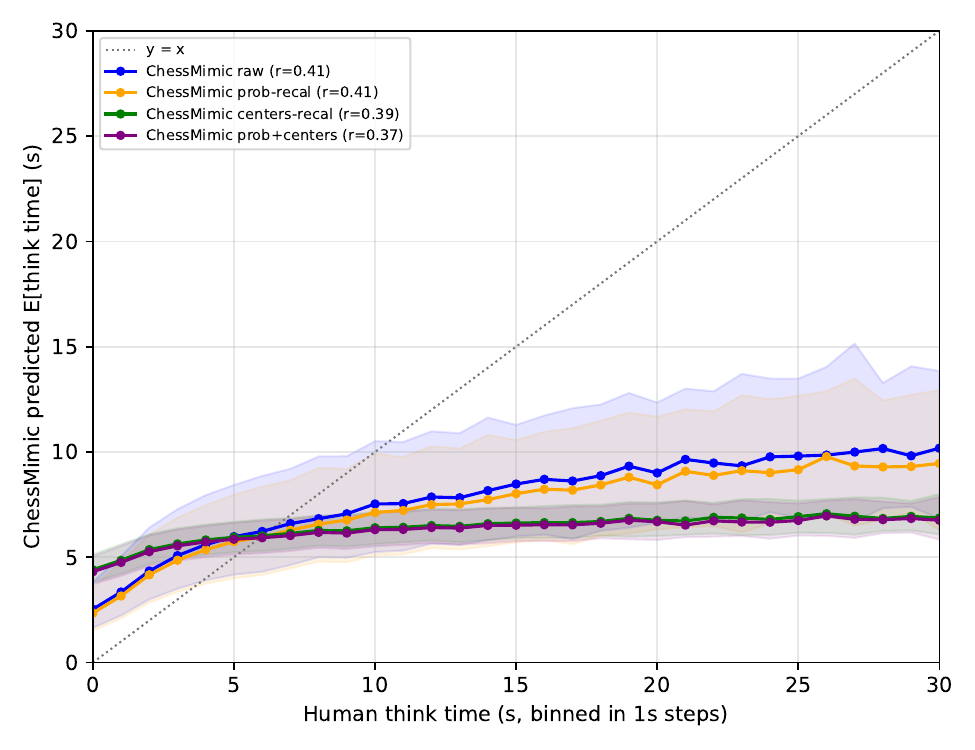}
\caption{Per-ply think-time prediction: median + IQR of
  $E[t]$ inside 1-second bins of human time, $n=89{,}299$. ALLIE-style
  reporting (\cite{zhang2025allie}, Figure 3). Four variants overlaid:
  raw clock-model output (blue, $r{=}0.41$), per-band probability
  label-shift correction (orange, $r{=}0.41$), per-band soft-binned
  center rescaling (green, $r{=}0.39$), and both fixes combined
  (purple, $r{=}0.37$). All four saturate well below $y{=}x$ at high
  human times; the centers and combined variants sit \emph{below} the
  raw curve because the calibrated centers collapse toward the
  population mean. The fact that neither post-hoc fix improves the
  correlation places the residual gap on per-position distribution
  uniformity rather than on bucket-level miscalibration.}
\label{fig:think_time}
\end{figure}

Figure~\ref{fig:think_time} overlays the ALLIE-style median + IQR plot
for the raw output and the three post-hoc variants. The prob-recal
curve sits modestly \emph{above} the raw one; the centers-recal and
combined curves sit \emph{below} it at high human times -- the visual
signature of the center collapse described in (i). All four curves
saturate well short of $y{=}x$ at high human times, visualizing the
conclusion that the saturating behavior is produced by per-position
distribution uniformity, not by bucket-level miscalibration.

The \emph{move} head's softmax is similarly diffuse on natural blitz,
and there we can test whether it is structural by changing the input
distribution. On Lichess tactical puzzles -- positions with a
uniquely-correct move -- the move head's softmax concentrates
meaningfully more than on band-matched natural blitz
(\S\ref{sec:tactics}; Appendix~\ref{app:interp},
\S\ref{app:concentration}), though it never puts all of its mass on one
move even on saturated motifs. So at least on the move-head side this
diffuseness is partly an input property rather than a structural
ceiling. We
have not run the analogous tactical-distribution test for the
clock head itself; the diagnostic above is the most we say about
clock-distribution uniformity directly.

\paragraph{(D) Blunder rate vs.\ remaining clock for human players.}
We label each played move by running Stockfish 17 at depth 12 and classifying it as \emph{ok}
($\text{cp\_loss} < 100$), \emph{mistake} ($100\!-\!200$), or
\emph{blunder} ($\geq 200$). Binning the resulting labels by the moving
side's remaining clock at the time of the move
(Table~\ref{tab:blunder_by_clock}) reveals two patterns. First, blunder
rates fall steadily as the clock grows -- $8.5\%$ with $>\!120$\,s
remaining vs $15.0\%$ below $5$\,s -- and at the month-wide sub-sample
size the trend is now monotone, with tight Wilson intervals
($15.0\% \to 13.1\% \to 13.0\% \to 12.2\% \to 11.8\% \to 8.5\%$). Second,
the \emph{magnitude} of the worst moves is larger under time pressure:
mean cp-loss is $384$\,cp at $>\!120$\,s and $1{,}710$\,cp at $<\!5$\,s,
broadly decreasing as the clock grows (with a small non-monotonicity
among the mid-clock buckets). The mistake-or-worse rate ($\geq\!100$\,cp)
is much flatter ($18.1\text{--}21.3\%$), so the time-pressure effect is
concentrated in the high-severity tail rather than in the small-error
tail. The specific cp-loss values should be read as approximate due to
the limited search depth, but the directional clock trend only relies on
the relative ordering of the cp-losses. This Stockfish-labelled
sub-sample was computed before the bot-account exclusion described in
\S\ref{sec:data} was rolled out; re-running it on the bot-filtered
sub-sample would remove $\approx\!0.3\%$ of positions distributed
across all clock buckets and would not change the qualitative
clock-vs-blunder trend.

\begin{table}[h]
\centering\small
\begin{tabular}{lrcccr}
\toprule
Remaining clock & $n$ & Blunder rate ($\geq\!200$\,cp) & 95\% CI & Mistake-or-worse ($\geq\!100$\,cp) & Mean cp-loss \\
\midrule
$<5$\,s     &     655 & \textbf{0.150} & 0.124--0.179 & 0.212 & 1{,}710 \\
$5\text{--}15$\,s   &   2{,}633 & 0.131 & 0.119--0.145 & 0.208 & 1{,}561 \\
$15\text{--}30$\,s  &   3{,}109 & 0.130 & 0.119--0.142 & 0.208 & 1{,}258 \\
$30\text{--}60$\,s  &   5{,}940 & 0.122 & 0.114--0.131 & 0.210 & 1{,}270 \\
$60\text{--}120$\,s & 17{,}165 & 0.118 & 0.113--0.123 & 0.213 & 943 \\
$>120$\,s   & 70{,}500 & \textbf{0.085} & 0.083--0.087 & 0.181 & 384 \\
\bottomrule
\end{tabular}
\caption{Human players' empirical move quality vs.\ moving-side's remaining clock on the
  $100{,}000$-position stratified sub-sample, labeled with Stockfish~17 at
  depth 12 (30-second wall-clock cap per call; depth locked by the tuning
  probe in \S\ref{sec:limitations}). Blunder rate is nearly double
  under severe time pressure ($<\!5$\,s) than at $>\!120$\,s and is
  monotone in the clock. Wilson $95\%$ intervals shown.}
\label{tab:blunder_by_clock}
\end{table}

\paragraph{(E) The move model under counterfactual clock.}
As we change the moving side's clock from $120$\,s to $1$\,s the move distribution
does not change dramatically - the mean Jensen-Shannon divergence between
the two legal-move distributions is only $\approx 0.04$ bits and the
top-1 move changes in just $16\text{--}21\%$ of positions. However, the 
probability distribution over legal moves becomes more diffuse. The top-1 
probability falls from $0.55$ to $0.47$ and the redistributed mass goes to worse moves. 
Weighting each legal move's Stockfish cp-loss (capped at $1{,}000$\,cp) by the
model's probability of playing it, the overall expected cp-loss 
rises from $88$\,cp at $120$\,s to $116$\,cp at $1$\,s. This happens
regardless of rating, and the cp-loss is positive in all 14 bands
(Table~\ref{tab:expected_cploss}). The model has learned that human play
degrades under time pressure. 

\begin{table}[h]
\centering\small
\begin{tabular}{l r r r l}
\toprule
Side-to-move band & $n$ & $c{=}120$\,s & $c{=}1$\,s & $\Delta$ (95\% CI) \\
\midrule
0-1000    &   200 & 131 & 172 & $+41.9$~[34.4, 49.6] \\
1000-1100 &   200 & 103 & 135 & $+31.4$~[25.6, 37.7] \\
1100-1200 &   200 &  98 & 132 & $+33.6$~[27.7, 40.1] \\
1200-1300 &   200 &  94 & 121 & $+27.1$~[22.3, 32.2] \\
1300-1400 &   200 &  92 & 121 & $+29.2$~[23.6, 35.1] \\
1400-1500 &   200 &  99 & 131 & $+32.4$~[26.4, 38.9] \\
1500-1600 &   200 &  74 & 103 & $+29.0$~[23.5, 34.8] \\
1600-1700 &   200 &  80 & 102 & $+22.6$~[18.0, 27.5] \\
1700-1800 &   200 &  92 & 121 & $+29.2$~[23.3, 35.8] \\
1800-1900 &   200 &  77 & 100 & $+23.3$~[19.0, 28.2] \\
1900-2000 &   200 &  84 & 105 & $+20.9$~[15.7, 26.1] \\
2000-2100 &   200 &  75 &  95 & $+20.7$~[14.6, 26.9] \\
2100-2200 &   200 &  76 & 100 & $+24.2$~[19.5, 29.5] \\
2200-3500 &   200 &  60 &  83 & $+22.5$~[17.8, 27.6] \\
\midrule
Overall   & 2{,}800 &  88 & 116 & $+27.7$~[26.2, 29.3] \\
\bottomrule
\end{tabular}
\caption{Probability-weighted expected centipawn loss of the move
  model's legal-move distribution under a counterfactual comfortable
  clock ($c{=}120$\,s) vs.\ severe time pressure ($c{=}1$\,s), on a
  $2{,}800$-position stratified sub-sample (200 per side-to-move band) of
  the April 2026 held-out set. The $c$ columns are
  $\mathbb{E}[\text{cp-loss}\mid c]=\sum_m P(m\mid c)\,\text{cp-loss}(m)$
  in cp, with each legal move's Stockfish~17 cp-loss (MultiPV, depth 12)
  capped at $1{,}000$ to limit the influence of rare mate-losing tail
  moves. $\Delta$ is the paired per-position increase with a bootstrap
  95\% CI. Expected move quality degrades under time pressure in every band.}
\label{tab:expected_cploss}
\end{table}

\subsection{Tactical accuracy on Lichess puzzles}
\label{sec:tactics}

So far the held-out evaluation has been on natural blitz, where each
position is a snapshot from a real game and the ``right'' answer is
the move a human actually played. To test \textsc{ChessMimic} on a
different distribution -- positions with a \emph{uniquely-correct}
move rather than an opinion -- we evaluate on the public Lichess
tactical puzzles database~\cite{lichess_database}: $\sim$5.94\,M
Stockfish-curated puzzles, each annotated with a Lichess puzzle
rating (a Glicko-2 estimate updated from user solving attempts, not
an engine evaluation) and a list of motif tags (\texttt{pin},
\texttt{fork}, \texttt{mateIn1..5}, \texttt{discoveredAttack},
\texttt{intermezzo}, \texttt{xRayAttack}, \texttt{attraction},
\texttt{deflection}, \texttt{sacrifice}, $\ldots$). We sample a
$1.25\%$ reservoir: $n=74{,}424$ puzzles with rating range
$399$--$3196$ (median $1420$). For each puzzle we replay the
Lichess ``setup move'' onto the recorded FEN to
obtain the solver position. Each
puzzle is scored once per band with \texttt{elo\_self} set to the
band midpoint (Maia convention -- the puzzle's Lichess rating
labels difficulty, not player skill) and a fixed comfortable clock
(120\,s, well above any time-pressure regime the clock model reacts
to). We report two Maia-2 baselines: \emph{(P-routed)} the Maia /
Allie convention where \texttt{elo\_self} is each puzzle's own
Lichess rating (a Glicko-2 estimate from user solving attempts, not
an engine evaluation), so the Maia-2 number is independent of the
\textsc{ChessMimic} band; and \emph{(M-routed)} a matched-routing
variant where Maia-2 is rerun once per band with \texttt{elo\_self}
set to that same band's midpoint, so the routing matches
\textsc{ChessMimic}'s.

\begin{table}[h]
\centering\small
\begin{tabular}{l r r r r}
\toprule
\textsc{ChessMimic} band & CM top-1 & Maia-2 (P) & Maia-2 (M) & $\Delta$ vs.\ M \\
\midrule
0-1000    & \textbf{0.526} & 0.655 & 0.637 & $-0.111$ \\
1000-1100 &        0.653  & 0.655 & 0.637 & $+0.016$ \\
1100-1200 &        0.662  & 0.655 & 0.648 & $+0.014$ \\
1200-1300 &        0.674  & 0.655 & 0.651 & $+0.023$ \\
1300-1400 &        0.682  & 0.655 & 0.653 & $+0.029$ \\
1400-1500 &        0.689  & 0.655 & 0.656 & $+0.033$ \\
1500-1600 &        0.696  & 0.655 & 0.658 & $+0.038$ \\
1600-1700 &        0.704  & 0.655 & 0.661 & $+0.043$ \\
1700-1800 &        0.711  & 0.655 & 0.661 & $+0.050$ \\
1800-1900 &        0.721  & 0.655 & 0.660 & $+0.061$ \\
1900-2000 &        0.730  & 0.655 & 0.661 & $+0.069$ \\
2000-2100 &        0.738  & 0.655 & 0.661 & $+0.077$ \\
2100-2200 &        0.747  & 0.655 & 0.661 & $+0.086$ \\
2200-3500 & \textbf{0.765} & 0.655 & 0.661 & $+0.104$ \\
\bottomrule
\end{tabular}
\caption{Per-band top-1 tactical accuracy on a fixed set of
  $n=74{,}424$ Lichess puzzles. Each \textsc{ChessMimic} row is the
  corresponding band's move model run on those puzzles with
  \texttt{elo\_self} fixed to the band midpoint. The \emph{Maia-2 (P)}
  column is a single Maia-2 run with \texttt{elo\_self}~$=$~each
  puzzle's Lichess rating (constant across rows by construction);
  \emph{Maia-2 (M)} is a matched-routing Maia-2 rerun for each
  band's midpoint, so the routing matches \textsc{ChessMimic}'s on
  that row. The matched-routing Maia-2 plateaus at $\approx 0.66$
  above band $1500$--$1600$, because Maia-2's internal Elo
  conditioning saturates: very low or very high midpoints get
  clamped to the trained range. Either baseline gives the same
  qualitative story: \textsc{ChessMimic} matches Maia-2 by band
  $1000$--$1100$ and beats it by $\geq\!10$\,pp at the top band.
  The low top-1 at the $0$--$1000$ band has two possible
  explanations from this metric: a $0$--$1000$ player would itself
  often miss high-rated tactics, so a model that faithfully imitates
  such a player should also miss them; and high-rated tactical
  positions are rare in $<\!1000$ blitz training data, so they are
  also out-of-distribution inputs for the $0$--$1000$ model.}
\label{tab:puzzles_per_band}
\end{table}

\textsc{ChessMimic}'s overall top-1 climbs monotonically with band,
from $0.526$ at band 0-1000 to $0.765$ at band 2200-3500. The gap
with Maia-2 grows similarly, from -0.111 to +0.104. Note that higher
is not necessarily better in these metrics, because Lichess Puzzle
ratings are not comparable to the player ratings that Maia and 
\textsc{ChessMimic} were trained on.
The full per-theme table is in Appendix~\ref{app:puzzles};
Table~\ref{tab:puzzle_headlines} pulls out three headline patterns --- the
lookahead-heavy themes whose accuracy climbs most steeply with rating, the
themes where \textsc{ChessMimic}'s top-band advantage over Maia-2 is
largest, and the long-horizon motifs it remains weakest on at the top band.

\begin{table}[h]
\centering\small
\begin{tabular}{l r l}
\toprule
Theme & \multicolumn{2}{l}{Top-1} \\
\midrule
\multicolumn{3}{l}{\emph{Steepest climb across bands} ($\Delta$, 0-1000 $\to$ 2200-3500)} \\
\texttt{intermezzo}       & $+0.335$ & ($0.297\!\to\!0.632$) \\
\texttt{xRayAttack}       & $+0.322$ & ($0.437\!\to\!0.759$) \\
\texttt{pin}              & $+0.182$ & \\
\texttt{attraction}       & $+0.171$ & \\
\texttt{sacrifice}        & $+0.149$ & \\
\midrule
\multicolumn{3}{l}{\emph{Largest advantage over Maia-2 at 2200-3500} ($\Delta$)} \\
\texttt{xRayAttack}       & $+0.332$ & ($0.759$ vs $0.427$) \\
\texttt{discoveredAttack} & $+0.194$ & \\
\texttt{intermezzo}       & $+0.190$ & \\
\texttt{attraction}       & $+0.175$ & \\
\texttt{deflection}       & $+0.153$ & \\
\texttt{fork}             & $+0.147$ & \\
\texttt{sacrifice}        & $+0.146$ & \\
\midrule
\multicolumn{3}{l}{\emph{Weakest themes at 2200-3500} (lowest top-1)} \\
\texttt{sacrifice}        & $0.432$ & \\
\texttt{attraction}       & $0.486$ & \\
\texttt{clearance}        & $0.529$ & \\
\texttt{mateIn4}          & $0.560$ & \\
\bottomrule
\end{tabular}
\caption{Headline per-theme patterns on the Lichess puzzle benchmark
  (full per-theme table in Appendix~\ref{app:puzzles}). \emph{Top}: themes
  whose top-1 accuracy climbs most from the 0-1000 to the 2200-3500 band.
  \emph{Middle}: themes with the largest top-1 advantage over Maia-2 at the
  2200-3500 band. \emph{Bottom}: the themes \textsc{ChessMimic} is weakest
  on at 2200-3500. Parentheses give the band-endpoint accuracies (top group)
  and the \textsc{ChessMimic}-vs-Maia-2 top-1 (middle group) for the leading
  theme in each.}
\label{tab:puzzle_headlines}
\end{table}

\paragraph{Move-head distribution concentration on puzzles, an
analogue of \S\ref{sec:clock_studies}(C).} The clock-head
diagnostic in \S\ref{sec:clock_studies}(C) found that the
\emph{clock} model's per-position softmax is close to its
population marginal on natural blitz. The move head's legal-move
softmax is likewise diffuse on natural blitz; we can test whether that
diffuseness is structural by switching the input distribution to
Lichess tactical puzzles (which have a uniquely-correct move) and
asking how concentrated \textsc{ChessMimic}'s legal-move softmax
is, compared to band-matched natural blitz. We measure normalized
entropy
$H_{\text{norm}} = -\sum p_i \log p_i / \log n_{\text{legal}}$,
where $0$ is all probability on one move (a one-hot
distribution) and $1$ is uniform on all legal moves. Across $12$
of $14$ bands (1000--2100), puzzles concentrate \emph{more} than
blitz: $\Delta H_{\text{norm}}$ widens monotonically from
$-0.024$ at band $1000$--$1100$ to $-0.061$ at band
$2000$--$2100$, then closes back to $-0.008$ at the top
$2200$--$3500$ band (where the measured entropy gap between puzzles
and natural blitz is small). The move head \emph{can} sharpen when the
position has a uniquely-correct move -- its diffuse
behaviour on natural blitz is therefore (at least partly) an
input property rather than a structural ceiling -- though even on
saturated motifs (\texttt{mateIn1}, \texttt{oneMove}) puzzle
$H_{\text{norm}}$ hovers at $\approx 0.20$, not $0$. A side-finding: per-theme mean
$H_{\text{norm}}$ tracks per-theme accuracy negatively
(at band $2200$+: \texttt{backRankMate} $H\!=\!0.20$, accuracy
$0.93$; \texttt{sacrifice} $H\!=\!0.46$, accuracy $0.43$), making
per-position softmax entropy a usable serve-time confidence
score. The full per-band curve and per-theme breakdown are in
Appendix~\ref{app:concentration}.

\subsection{Exploratory attention diagnostics on the move encoder}
\label{sec:interp}

In this exploratory section we treat last-layer head-averaged CLS
attention as a soft proxy for ``what part of the input the model
weights when scoring a position'' and look for descriptive patterns
that correlate with prediction quality. We do not run causal
interventions (e.g. attention patching, ablation, or probe-based controls),
so the findings should be read as qualitative observations and not
mechanistic claims. The attention interpretability literature
\cite{jain2019attention,wiegreffe2019attention} cautions against
stronger readings. Mechanistic interpretability of transformers has
been a research focus area in language models, both at the head
level \cite{voita2019heads,clark2019bert,michel2019heads} and in the
``circuits'' line of work \cite{elhage2021framework}. The
closest precedent in the chess domain is McGrath et~al.'s linear-probe
analysis of AlphaZero \cite{mcgrath2022alphazero}, which finds that
the network recovers human chess concepts during training. We
report two diagnostic findings on the move encoder at band
$2200$+: an attention region pattern across the encoder layers, and
an attention centered failure typology on confidently wrong puzzles.

\paragraph{Per-layer attention region pattern.}
Table~\ref{tab:layer_pipeline} shows how attention to the board,
recent moves, conditioning (e.g. rating, clock, side to move),
and other tokens (e.g. castling, en-passant, move number) varies as
data flows from early layers to later layers. We observe that
the model is board-focused in early and late layers, but diffuse and
conditioning focused in middle layers. This is reminiscent of head
specialization patterns documented in NMT and BERT \cite{voita2019heads,clark2019bert}:
an early per-layer ``ingest the board'' phase, three intermediate 
layers that integrate the conditioning tokens, and a late re-localization
back onto the board for the policy head. Among the 64 individual (layer, head) units,
three have a particularly interpretable focus. Layer 0 head 7 focuses on recent move
tracking, Layer 4 head 3 focuses on rating and clock, and layer 6 head 3 focuses on the a8-h1
diagonal. Seven other heads focus on specific areas of the board but are more diffuse. For
example, we find heads focused on fianchetto squares, castled king destination, central squares,
back ranks, and others. All are catalogued in Appendix~\ref{app:heads} (Table~\ref{tab:named_heads}) 
along with the full clustering analysis, PCA scatter, and per-head heatmap grid.

\begin{table}[h]
\centering\small
\begin{tabular}{c r r r r}
\toprule
Layer & board & recent & cond & other \\
\midrule
0 & 0.57 & 0.13 & 0.10 & 0.21 \\
1 & 0.57 & 0.08 & 0.17 & 0.18 \\
2 & 0.29 & 0.07 & 0.31 & 0.33 \\
3 & 0.31 & 0.11 & 0.31 & 0.27 \\
4 & 0.43 & 0.09 & 0.28 & 0.20 \\
5 & 0.65 & 0.06 & 0.19 & 0.10 \\
6 & 0.84 & 0.03 & 0.11 & 0.03 \\
7 & 0.89 & 0.02 & 0.08 & 0.01 \\
\bottomrule
\end{tabular}
\caption{Mean fraction of CLS-token attention on the four input
  regions, per encoder layer (band $2200$+, head-averaged,
  $7{,}651$ puzzles across $17$ themes). Row entries sum to $1$ by
  softmax invariance. The board mass dips in layers $2$--$4$
  (replaced mostly by \texttt{cond} and \texttt{other}) and
  re-concentrates from layer $5$ onward, peaking at $0.89$ at the
  policy-output layer.}
\label{tab:layer_pipeline}
\end{table}

\paragraph{Failure typology on confidently-wrong puzzles.}
We further dissect \emph{which} puzzles the strongest band
($2200$+) gets wrong with high confidence by inspecting where its
CLS attention landed. From the Lichess puzzles set we filter to
puzzles with rating $\geq 1800$, $\geq 8$ legal moves,
\textsc{ChessMimic} top-1 mass $\geq 0.5$, and top-1
$\neq$ played move. $8{,}540$ puzzles meet the criteria, and of these we
sub-sample to 800 for more detailed analysis.  For each
puzzle we compute three quantities on the head-averaged last layer
attention: \texttt{key\_mass}, the attention on the target move's
\{from, to\} squares; \texttt{predicted\_mass}, the same on the
model's wrong-pick endpoints; and \texttt{recent\_mass}, the sum over
the 12 recent-move tokens. The typology
(Table~\ref{tab:failure_typology}) sorts each puzzle into one of
three failure modes. The classification is not exhaustive - 
$427$ of the $800$ confidently-wrong puzzles ($53.4\%$) fall outside 
all three rules and are left unclassified. Among
the $373$ classified failures, attention misses (Type A, $n=254$)
outnumber downstream-policy errors (Type B, $n=39$) by
$\sim 6.5\times$. Within the classified subset, the errors
correlate more with low attention on the target move's endpoints
than with policy-head confusion given correct attention. We report
this as a correlation, not a causal claim that the policy head
deterministically follows attention. Three case-study walkthroughs,
one per type, are in Appendix~\ref{app:failures}.

\begin{table}[h]
\centering\small
\begin{tabular}{l l r r}
\toprule
Type & Rule & $n$ & \% of $800$ \\
\midrule
A (attention miss)
  & \texttt{key\_mass} $<\!0.07$ and \texttt{predicted\_mass} $>$ \texttt{key\_mass}
  & 254 & $31.8\%$ \\
B (right attention, wrong move)
  & \texttt{key\_mass} $\geq\!0.15$ & 39  & $4.9\%$ \\
C (relative recent-mass tail)
  & \texttt{recent\_mass} $\geq$ pool $p_{90}$ & 51  & $6.4\%$ \\
mixed-AC & A and C & 20  & $2.5\%$ \\
mixed-BC & B and C &  9  & $1.1\%$ \\
unclassified & none of the above & 427 & $53.4\%$ \\
\bottomrule
\end{tabular}
\caption{Failure-mode counts on the $n=800$ confidently-wrong
  sub-sample at band $2200$+. Type-A thresholds anchored to the
  Appendix~B.3 mean incorrect \texttt{key\_mass} ($0.087$); Type-B
  threshold above the mean correct ($0.140$). The absolute
  \texttt{recent\_mass} never reaches the $0.10$ value
  Appendix~B.4's layer-7 head-averaged baseline ($\approx 0.03$)
  would have implied as ``fixation'', so Type C is defined as the
  pool's $p_{90}$ -- the relative tail of recent-move attention
  within confidently-wrong puzzles. \textbf{Type A outnumbers
  Type B by $\approx\!6.5\times$.}}
\label{tab:failure_typology}
\end{table}

\section{Deployment}
\label{sec:deployment}

\textsc{ChessMimic} runs as the backend of the public demo at
\href{https://1e4.ai}{1e4.ai}. This section describes the serving path,
its memory footprint, and its measured latency.

\paragraph{Common moves database.}
Positions skipped during the data build because they were too common
(\S\ref{sec:data}) are not included in the move model's training data,
and therefore are not learned by the move model. Instead, the empirical
move distribution for each skipped \texttt{(FEN, last-12-moves)} key is
stored in a common moves database for each band. At serve time, a request first
hits this database. If the key is in the database, a move is sampled empirically
from the stored human distribution. Positions not in the database fall through to 
the move model.
The benchmark in \S\ref{sec:experiments} disables this lookup so that the reported
\textsc{ChessMimic} numbers reflect the move model alone. The database's measured
effect on held-out accuracy (\S\ref{sec:movepred}) is negligible. The
database hits only $3.2\%$ of held-out positions and changes overall
top-1 by $-0.02$\,pp, so it is a latency and sampling optimization rather
than a source of the accuracy gap.

\paragraph{Serving footprint.}
Inference-time parameter counts and weight sizes are in
Table~\ref{tab:footprint}. Each task model is $\approx 9$M parameters
(38\,MB fp32, 19\,MB bf16); the deployed system holds $14\times 3 = 42$
checkpoints. 

\begin{table}[h]
\centering\small
\begin{tabular}{l r r r r}
\toprule
Model & Params & fp32 size & bf16 size & Checkpoints \\
\midrule
Move   & $9.45$M & $37.8$\,MB & $18.9$\,MB & 14 \\
Clock  & $8.95$M & $35.8$\,MB & $17.9$\,MB & 14 \\
Winner & $8.94$M & $35.8$\,MB & $17.9$\,MB & 14 \\
\midrule
Total deployed & $\approx 383$M & $\approx 1.5$\,GB & $\approx 0.8$\,GB & 42 \\
\bottomrule
\end{tabular}
\caption{Inference-time parameter counts and weight sizes (band 1500-1600
  representative; other bands are within $0.1\%$).}
\label{tab:footprint}
\end{table}

\paragraph{Serving latency.}
Table~\ref{tab:latency} reports single-position latency measured on
CPU (the deployment default) for the band-1500-1600 checkpoints, over
200 timed calls after a 20-call warmup.
Each path is single-digit milliseconds; a served bot move is
either a database hit ($\sim$8\,$\mu$s, effectively free) or, for tail
positions, the move model plus -- where the UI needs them -- the clock
and winner models, summing to $\sim$20\,ms median. This is comfortably
within interactive latency for a web app and is why the deployment runs
on commodity CPU without GPUs.

\begin{table}[h]
\centering\small
\begin{tabular}{l r r}
\toprule
Serving path & Median (ms) & p95 (ms) \\
\midrule
Common moves DB lookup            & $<0.01$ & $<0.01$ \\
Move model                        & 7.2  & 7.9 \\
Clock model                       & 6.4  & 7.0 \\
Winner model                      & 6.2  & 7.0 \\
\midrule
Neural triple (move+clock+winner) & 19.9 & 21.8 \\
\bottomrule
\end{tabular}
\caption{CPU serving latency, band 1500-1600, single-position calls
  ($n{=}200$ timed, 20 warmup). The common moves DB lookup is a hash
  lookup ($\sim$8\,$\mu$s). The three-model row is the worst case for a
  tail position when the UI requests move, predicted think-time, and the
  win-probability bar together.}
\label{tab:latency}
\end{table}

\section{Limitations}
\label{sec:limitations}

\paragraph{Controlled retraining ablations.}
We do not train a unified backbone \textsc{ChessMimic} variant, a 
clock-free winner model, a shared-backbone multihead variant, or a 
move model with a cross-entropy loss. Each one would require retraining 
a substantial fraction of the 42 checkpoints that comprise the deployed 
ensemble. The released \textsc{ChessMimic} configuration outperforms the 
released Maia-2 and Maia-3-5M baselines on the held-out slice, but we do 
not isolate the marginal contribution of our design choices (e.g.
per-band specialization, loss choice, architecture, clock features, or
the fine-tuning cascade).

We focus on blitz; bullet and classical time controls have different
time-usage statistics and would require new clock buckets and new
common-position thresholds at minimum.

The more expensive counterfactual studies
(\S\ref{sec:clock_studies}) run on a $100{,}000$-position
per-band-stratified sub-sample; their Stockfish blunder labels use a
depth-12 search with a $30$-second wall-clock cap, the depth chosen by a
calibration probe so the full sub-sample finishes within a fixed compute
budget, so the cp-loss \emph{magnitudes} there are approximate even
though the directional clock trends are tight.

We do not compare against the Maia-2 \texttt{rapid} or
\texttt{bullet} checkpoints, since the targeted regime is mismatched.
Our Maia-3 comparison (\S\ref{sec:chessformer_compare}) is now a direct
head-to-head: we run Maia-3-5M / 23M / 79M on our $849{,}730$-position
reconstruction of the ALLIE 2022 slice, and all three sizes reproduce
the reported top-1 to within $0.25$\,pp (a second validation point
alongside the $0.2$\,pp Maia-2 anchor). The residual is the $4\%$
slice-reconstruction gap, which appears to be marginally easier than
the full $884{,}049$-position set across all four anchor models we can
score.

We also do not run a search component on top of the policy -- ALLIE's
time-adaptive MCTS work~\cite{zhang2025allie} is a clear improvement path
which we have not attempted. Chessformer's Geometric Attention
Bias~\cite{monroe2026chessformer} is another concrete architectural
upgrade we have not yet evaluated, and one which our correspondence with
its authors suggests may explain a meaningful fraction of the cross-model
top-1 gap once data and parameter counts are held constant. Our per-ply
think-time Pearson $r$ ($0.410$, \S\ref{sec:clock_studies} (C)) is well
below ALLIE's reported $0.70$; we attribute most of the gap to the
$K{=}30$ bucketed clock head (vs.\ ALLIE's scalar regression head).

\section{Conclusion}
\label{sec:conclusion}

\textsc{ChessMimic} demonstrates that a deliberately simple recipe
can substantially outperform Maia-2 and Maia-3-5M on move prediction
and particularly on outcome calibration. \textsc{ChessMimic} uses
standard learned positional encoding without chess-specific specializations
and the architecture is a simple set of transformer layers that
have become standard in LLM implementations.
\textsc{ChessMimic}'s large advantage in win prediction appears to come from using a
separate 3-class (W/L/D) winner model with clock inputs.
We measure this empirically and do not claim it as a design failure of 
Maia-2, which was not optimized for clock-aware outcome forecasting. 
Our hope is that the code release and the explicit benchmark methodology 
make it easy for follow-up work to explore improvements and compare against
\textsc{ChessMimic}.

\paragraph{Acknowledgments.}
This work was conducted independently. We thank the Lichess team for
maintaining the open chess game database. The data pipeline reuses
\texttt{bagz} format code and the FEN tokenizer from Google DeepMind's
\texttt{searchless\_chess} repository \cite{ruoss2024searchless} under
Apache 2.0; the model design was informed by the Maia line of work
\cite{mcilroyyoung2020maia,tang2024maia2}. None of the cited authors are
affiliated with this work and they should not be assumed to endorse it.

\appendix

\section{Lichess puzzle benchmark: full per-theme accuracy}
\label{app:puzzles}

Table~\ref{tab:puzzles_per_theme} extends \S\ref{sec:tactics}'s
headline table with the per-theme breakdown for themes with more than 200 samples. 
Each row is one
Lichess theme tag (a puzzle can carry multiple tags; rows are
inclusive unions). The five model columns are \textsc{ChessMimic}'s
per-band top-1 at four representative bands plus the Maia-2
baseline (same per-puzzle routing as Table~\ref{tab:puzzles_per_band}).

\setlength{\LTcapwidth}{\textwidth}
\begin{longtable}{l r r r r r r}
\toprule
Theme & $n$ & CM 1000-1100 & CM 1500-1600 & CM 2100-2200 & CM 2200-3500 & Maia-2 \\
\midrule
\endfirsthead
\toprule
Theme & $n$ & CM 1000-1100 & CM 1500-1600 & CM 2100-2200 & CM 2200-3500 & Maia-2 \\
\midrule
\endhead
\bottomrule
\endfoot
\textbf{overall}      & 74{,}424 & 0.653 & 0.696 & 0.747 & 0.765 & 0.655 \\
short                 & 37{,}633 & 0.651 & 0.702 & 0.762 & 0.781 & 0.644 \\
endgame               & 37{,}264 & 0.677 & 0.708 & 0.751 & 0.771 & 0.657 \\
middlegame            & 33{,}564 & 0.626 & 0.681 & 0.738 & 0.755 & 0.649 \\
crushing              & 28{,}833 & 0.609 & 0.648 & 0.691 & 0.713 & 0.596 \\
mate                  & 23{,}704 & 0.763 & 0.803 & 0.854 & 0.868 & 0.779 \\
advantage             & 21{,}378 & 0.592 & 0.643 & 0.704 & 0.722 & 0.597 \\
long                  & 19{,}427 & 0.572 & 0.605 & 0.643 & 0.664 & 0.579 \\
oneMove               & 11{,}006 & 0.845 & 0.886 & 0.938 & 0.951 & 0.870 \\
mateIn1               & 10{,}975 & 0.845 & 0.887 & 0.939 & 0.951 & 0.871 \\
master                & 10{,}114 & 0.619 & 0.658 & 0.711 & 0.739 & 0.614 \\
mateIn2               &  9{,}892 & 0.727 & 0.766 & 0.816 & 0.829 & 0.731 \\
fork                  &  9{,}610 & 0.681 & 0.714 & 0.766 & 0.795 & 0.647 \\
kingsideAttack        &  6{,}526 & 0.637 & 0.704 & 0.760 & 0.772 & 0.696 \\
veryLong              &  6{,}000 & 0.579 & 0.601 & 0.634 & 0.652 & 0.576 \\
sacrifice             &  5{,}592 & 0.283 & 0.319 & 0.393 & 0.432 & 0.285 \\
advancedPawn          &  4{,}684 & 0.623 & 0.640 & 0.689 & 0.718 & 0.623 \\
pin                   &  4{,}615 & 0.504 & 0.578 & 0.660 & 0.686 & 0.550 \\
defensiveMove         &  4{,}502 & 0.645 & 0.669 & 0.695 & 0.707 & 0.644 \\
rookEndgame           &  4{,}045 & 0.637 & 0.661 & 0.685 & 0.722 & 0.607 \\
discoveredAttack      &  3{,}639 & 0.540 & 0.625 & 0.681 & 0.681 & 0.487 \\
opening               &  3{,}596 & 0.659 & 0.720 & 0.791 & 0.794 & 0.690 \\
deflection            &  3{,}287 & 0.505 & 0.536 & 0.613 & 0.645 & 0.493 \\
quietMove             &  3{,}112 & 0.514 & 0.560 & 0.618 & 0.652 & 0.543 \\
pawnEndgame           &  2{,}835 & 0.666 & 0.695 & 0.734 & 0.762 & 0.673 \\
hangingPiece          &  2{,}682 & 0.851 & 0.879 & 0.921 & 0.930 & 0.869 \\
attraction            &  2{,}648 & 0.315 & 0.353 & 0.434 & 0.486 & 0.311 \\
backRankMate          &  2{,}499 & 0.865 & 0.888 & 0.922 & 0.928 & 0.834 \\
mateIn3               &  2{,}428 & 0.586 & 0.624 & 0.677 & 0.698 & 0.610 \\
exposedKing           &  2{,}183 & 0.559 & 0.589 & 0.616 & 0.641 & 0.564 \\
promotion             &  1{,}840 & 0.653 & 0.671 & 0.739 & 0.766 & 0.649 \\
skewer                &  1{,}571 & 0.658 & 0.680 & 0.684 & 0.700 & 0.610 \\
discoveredCheck       &  1{,}320 & 0.530 & 0.570 & 0.642 & 0.650 & 0.547 \\
queensideAttack       &  1{,}175 & 0.689 & 0.706 & 0.751 & 0.769 & 0.676 \\
bishopEndgame         &  1{,}004 & 0.609 & 0.602 & 0.637 & 0.664 & 0.562 \\
clearance             &     949 & 0.490 & 0.522 & 0.519 & 0.529 & 0.518 \\
masterVsMaster        &     923 & 0.592 & 0.615 & 0.663 & 0.696 & 0.569 \\
intermezzo            &     875 & 0.297 & 0.438 & 0.587 & 0.632 & 0.442 \\
queenEndgame          &     864 & 0.597 & 0.644 & 0.686 & 0.704 & 0.578 \\
trappedPiece          &     825 & 0.486 & 0.542 & 0.601 & 0.636 & 0.533 \\
pillsburysMate        &     823 & 0.706 & 0.742 & 0.817 & 0.830 & 0.717 \\
operaMate             &     805 & 0.682 & 0.739 & 0.802 & 0.832 & 0.712 \\
zugzwang              &     805 & 0.677 & 0.698 & 0.720 & 0.733 & 0.676 \\
knightEndgame         &     617 & 0.665 & 0.708 & 0.734 & 0.765 & 0.622 \\
attackingF2F7         &     602 & 0.746 & 0.821 & 0.854 & 0.854 & 0.832 \\
queenRookEndgame      &     576 & 0.747 & 0.788 & 0.825 & 0.821 & 0.741 \\
capturingDefender     &     473 & 0.478 & 0.537 & 0.613 & 0.647 & 0.482 \\
doubleCheck           &     374 & 0.471 & 0.511 & 0.620 & 0.631 & 0.513 \\
mateIn4               &     339 & 0.484 & 0.513 & 0.537 & 0.560 & 0.507 \\
smotheredMate         &     298 & 0.725 & 0.779 & 0.879 & 0.893 & 0.782 \\
epauletteMate         &     294 & 0.663 & 0.731 & 0.810 & 0.840 & 0.718 \\
xRayAttack            &     286 & 0.437 & 0.479 & 0.661 & 0.759 & 0.427 \\
interference          &     273 & 0.571 & 0.612 & 0.652 & 0.659 & 0.579 \\
\caption{Per-band per-theme top-1 accuracy on the
  $74{,}424$-puzzle reservoir (themes with $n\!\geq\!200$ shown,
  sorted by descending $n$). \textbf{CM $X$--$Y$} columns are
  \textsc{ChessMimic} at band $X$--$Y$ with
  \texttt{elo\_self}~=~band midpoint; the Maia-2 column uses the
  puzzle's Lichess rating as \texttt{elo\_self}, so it is
  band-independent and identical across rows in the underlying
  evaluation. The steepest climbs across bands are on the
  multi-step / lookahead motifs (\texttt{intermezzo}
  $+0.335$, \texttt{xRayAttack} $+0.322$, \texttt{pin} $+0.182$,
  \texttt{attraction} $+0.171$, \texttt{sacrifice} $+0.149$); the
  saturated tails are \texttt{mateIn1}, \texttt{oneMove},
  \texttt{hangingPiece}, \texttt{backRankMate} (all $>\!0.85$
  already at band $1000$--$1100$); the hard tails are
  \texttt{sacrifice}, \texttt{attraction}, \texttt{clearance},
  \texttt{mateIn4} at $0.43$--$0.56$ even at band $2200$--$3500$.}
\label{tab:puzzles_per_theme}
\end{longtable}

\section{Attention diagnostics: supporting detail}
\label{app:interp}

\S\ref{sec:interp} carried the headline attention diagnostics (the
per-layer attention-region pattern and the failure-mode typology).
This
appendix carries the full machinery: setup, the per-position
distribution-concentration figure, the v1/v2 key-square analysis,
the head-clustering analysis and named specialists, the cross-band
attention trajectories, and the failure-mode case studies.

\subsection{Setup: capturing per-head attention}
\label{app:interp_setup}

The 92-token input sequence is laid out as: positions 0--11
recent-move tokens, 12 rating, 13 clock, 14 side-to-move, 15--78
board squares (64 squares in FEN order \texttt{a8} top-left through
\texttt{h1} bottom-right), 79--82 castling rights, 83--84
en-passant, 85--87 halfmove clock, 88--90 fullmove number, 91 CLS
(the only token projected to the policy output). All analyses
slice the CLS-token row of the last-layer attention and (except
where noted) average over the 8 heads, mirroring how the policy
head consumes it.

Two key-square definitions are used in the appendix sections:
\textbf{v1} = the $\{$from, to$\}$ of the target move (2 squares
per puzzle); \textbf{v2} = v1 plus the from / to squares of every
move in the Lichess solution PV plus motif-specific extensions
(post-move attacks for \texttt{fork} and \texttt{doubleCheck},
newly-unmasked king attackers for \texttt{discoveredAttack} and
\texttt{discoveredCheck}, the enemy king for mate-family themes).
v2 is a strict superset of v1.

\subsection{Per-position distribution concentration}
\label{app:concentration}

\begin{figure}[h]
\centering
\includegraphics[width=0.95\textwidth]{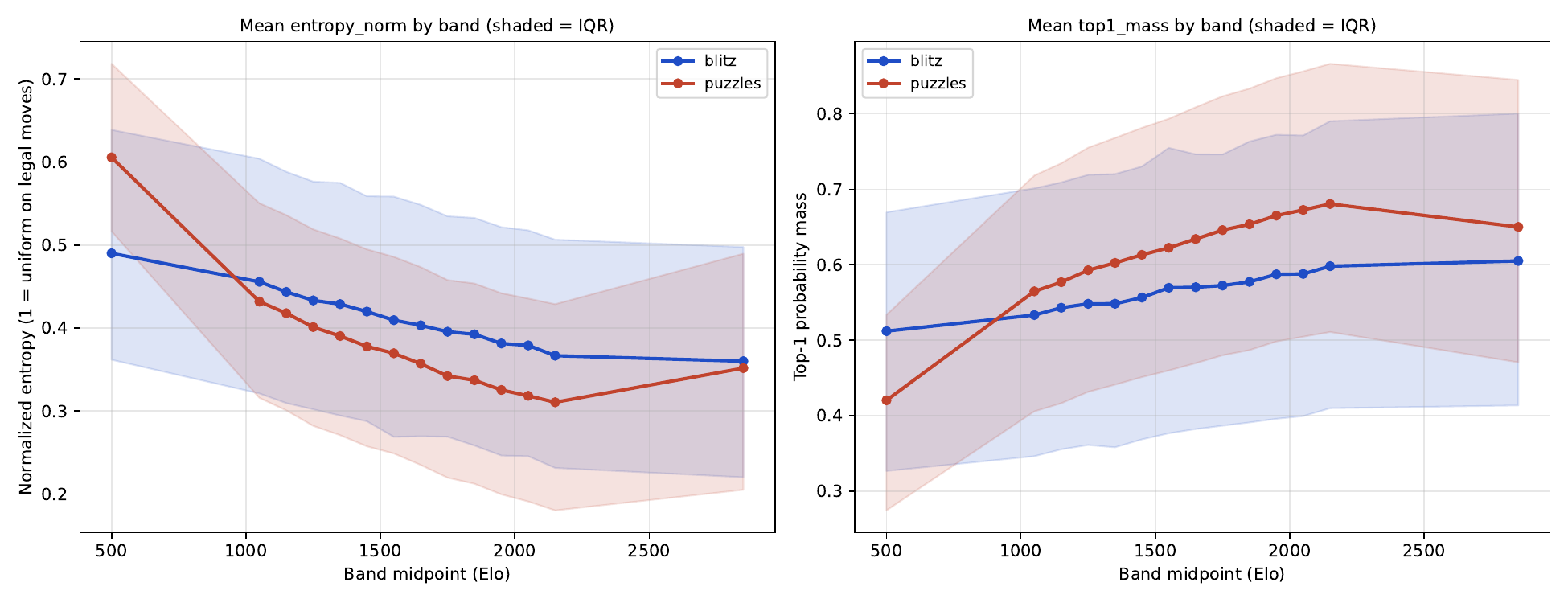}
\caption{Per-band mean normalized entropy (left) and top-$1$ mass
  (right) on the natural-blitz held-out set (blue, $n=7{,}143$ per
  band) vs the full Lichess puzzles set (red, $n=74{,}424$ scored
  $14$ times -- once per band -- with \texttt{elo\_self} =
  midpoint). IQR shaded. Puzzles concentrate \emph{more} than
  band-matched blitz across $12$ of $14$ bands; the gap widens
  monotonically until $2000$--$2100$, then closes at $2200$--$3500$
  where the measured entropy gap between puzzles and natural blitz is
  small. The $0$--$1000$
  band inverts: the weakest model sees high-rated puzzles as
  out-of-distribution and outputs near-uniform.}
\label{fig:entropy_per_band}
\end{figure}

Figure~\ref{fig:entropy_per_band} reports the per-band entropy and
top-$1$ mass comparison summarized in \S\ref{sec:tactics}. For
each of the 14 bands, the natural-blitz panel runs each position
through the band's model with the player's actual ELO as
\texttt{elo\_self}; the puzzles panel runs the full $74$\,k puzzles
through that band's model with \texttt{elo\_self} = band midpoint.

Two observations in the figure are worth pulling out: at the $0$--$1000$
band the sign of the entropy gap flips (puzzles $H_{\text{norm}} =
0.606$ vs blitz $0.490$); the band-$1000$ model trained on
sub-1000 blitz games sees high-difficulty puzzles as
out-of-distribution and expresses its uncertainty as high entropy. 
At band $2200$--$3500$ the gap between puzzle and natural blitz
nearly closes ($-0.008$); natural blitz here is already
near-optimal (super-GM positions) so the additional puzzle margin
is small. The $12$ middle bands show a monotone widening of
the gap with band, mirroring the per-band specialization story in
\S\ref{sec:tactics}.

Per-theme entropy at band $2200$+ tracks per-theme accuracy
negatively. The five themes with the lowest mean
$H_{\text{norm}}$ are also the five with the highest top-$1$
accuracy: \texttt{backRankMate} ($H\!=\!0.204$, $\text{acc}=0.928$),
\texttt{smotheredMate} ($0.214$, $0.893$),
\texttt{mateIn1} ($0.220$, $0.951$),
\texttt{oneMove} ($0.221$, $0.951$),
\texttt{hangingPiece} ($0.247$, $0.930$). The five themes with the
highest mean entropy include the ones the model is worst at:
\texttt{bishopEndgame} ($0.480$, $0.664$),
\texttt{clearance} ($0.457$, $0.529$),
\texttt{sacrifice} ($0.455$, $0.432$),
\texttt{defensiveMove} ($0.443$, $0.707$),
\texttt{trappedPiece} ($0.436$, $0.636$). 

\subsection{Key-square attention on tactical motifs}
\label{app:keysquares}

A 4{,}490-puzzle subset capped at 600 per theme across 9
motif-rich themes (\texttt{intermezzo}, \texttt{xRayAttack},
\texttt{pin}, \texttt{attraction}, \texttt{sacrifice},
\texttt{discoveredAttack}, \texttt{deflection}, \texttt{fork},
\texttt{mateIn1}) tests whether CLS attention concentrates on the
target move's endpoints (v1) or on the wider tactical net (v2).
Table~\ref{tab:keymass_overall} reports the overall mean
\texttt{key\_mass} for correct vs incorrect predictions at the
two endpoint bands. v1 cleanly separates correct from incorrect
predictions in every theme at both bands; v2 has higher absolute
mass but a narrower correct-vs-incorrect gap, suggesting the model
attends to the move it is about to play rather than to the broader
pattern.

\begin{table}[h]
\centering\small
\begin{tabular}{l l r r r r r r}
\toprule
Band & Variant & $n_c$ & km (correct) & km (incorrect) & $\Delta$ \\
\midrule
1200-1300 & v1 & 2{,}485 & 0.125 & 0.084 & $+0.041$ \\
1200-1300 & v2 & 2{,}485 & 0.201 & 0.201 & $-0.000$ \\
$2200$+    & v1 & 3{,}152 & 0.140 & 0.087 & $+0.053$ \\
$2200$+    & v2 & 3{,}152 & 0.228 & 0.211 & $+0.018$ \\
\bottomrule
\end{tabular}
\caption{Overall mean key-square attention (\texttt{key\_mass}) on
  the $n=4{,}490$ puzzles. With $2/64\approx 0.031$ as the uniform
  baseline on v1, the correct rows are $\approx\!4\times$ baseline
  and the incorrect rows $\approx\!2.5\times$. v2 widens the
  absolute value (more squares) but mostly narrows the
  correct-vs-incorrect gap -- the model attends to the endpoints
  it is about to play, not to the full Stockfish-PV tactical
  net. A paired per-puzzle cross-band shift (high $-$ low) is
  $+0.017$ on v1 and $+0.022$ on v2: higher-band models
  concentrate slightly more on the key squares.}
\label{tab:keymass_overall}
\end{table}

\begin{figure}[h]
\centering
\includegraphics[page=1,width=0.85\textwidth]{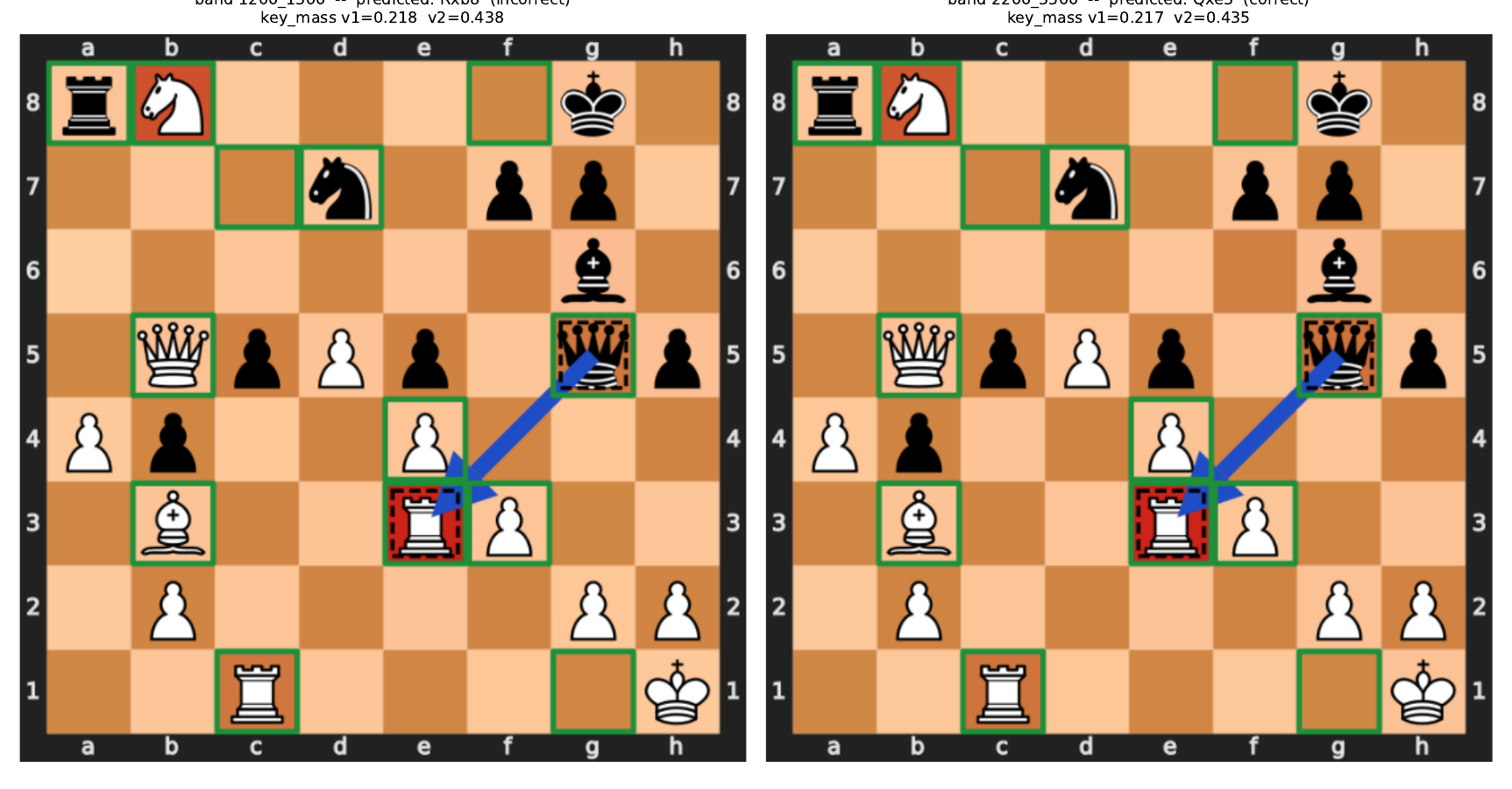}
\caption{Puzzle \texttt{0mjcb} (rating $2456$, themes
  \texttt{advantage defensiveMove fork hangingPiece intermezzo
  middlegame veryLong}, target \texttt{Qxe3}): low-band
  ($1200$--$1300$) and high-band ($2200$+) panels with the
  attention heatmap (red overlay), the solution-move arrow (blue),
  v1 key squares (dashed black, $\{$g5, e3$\}$), and v2 key
  squares (solid green, covering the wider fork/intermezzo net).
  At the low band the model predicts the tempting but losing
  \texttt{Rxb8}; at the high band the model correctly plays
  \texttt{Qxe3}. Both bands push attention to the move endpoints;
  the high band's attention is denser on the e3
  resolution square.}
\label{fig:atlas_0mjcb}
\end{figure}

\subsection{Per-head specialization}
\label{app:heads}

Capturing the CLS-to-full-sequence attention at every (layer,
head) on a $7{,}651$-puzzle subset (Phase 2's 9 themes plus
\texttt{kingsideAttack}, \texttt{queensideAttack},
\texttt{backRankMate}, \texttt{smotheredMate},
\texttt{doubleCheck}, \texttt{advancedPawn},
\texttt{defensiveMove}, \texttt{exposedKing}) gives a per-head
``global signature'' in $\mathbb{R}^{92}$. Running k-means with
silhouette over $k \in [2, 10]$ on the 64 signatures picks
$k\!=\!2$ in both bands. The split is even cleaner at the top
band (silhouette $0.493$ at $2200$+ vs $0.375$ at
$1200$--$1300$), and the two clusters split nearly identically
across bands:

\begin{table}[h]
\centering\small
\begin{tabular}{c r r r r r l}
\toprule
Cluster & $n$ & board & recent & cond & other & Where it lives \\
\midrule
1 (board-focused) & 40 & 0.72 & 0.06 & 0.12 & 0.10 & layers 0, 1, 4, 5, 6, 7 \\
0 (diffuse)        & 24 & 0.32 & 0.09 & 0.32 & 0.27 & layers 1--5 (mostly 2, 3) \\
\bottomrule
\end{tabular}
\caption{Two head clusters identified by k-means + silhouette on
  the 64 (layer, head) global signatures (band $2200$+).
  The split is essentially along PC1 of the signatures
  ($68.7\%$ variance at band $2200$+, per Figure~\ref{fig:head_clusters}),
  which is the board-mass axis. The diffuse
  cluster lives in the middle layers, where the per-layer
  pipeline (Table~\ref{tab:layer_pipeline}) integrates
  side-info before re-localizing onto the board.}
\label{tab:head_clusters}
\end{table}

\begin{figure}[h]
\centering
\includegraphics[page=2,width=\textwidth]{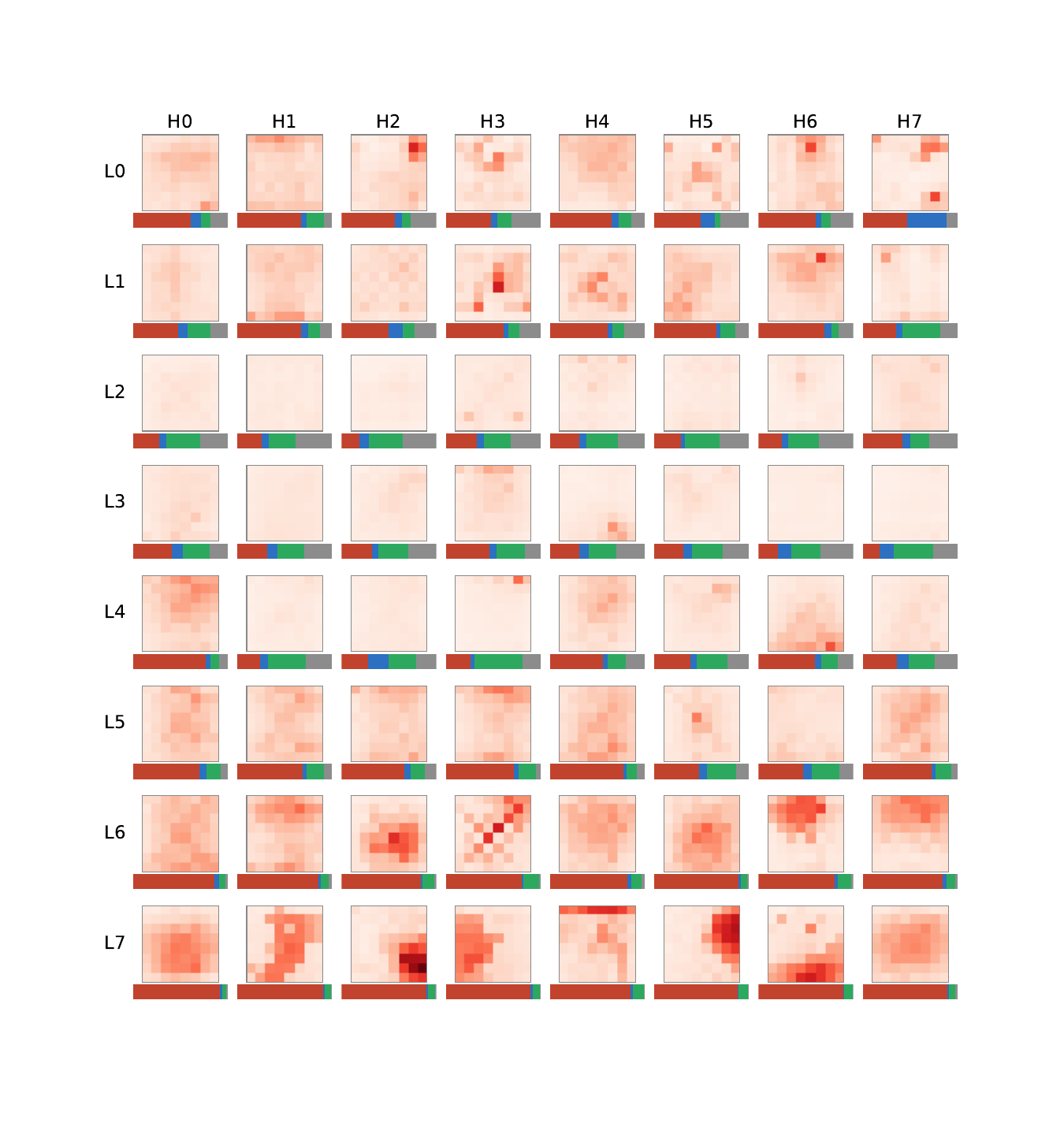}
\caption{Per-head global CLS-to-board attention at band
  $2200$+. Rows are layers
  $0$--$7$ (labelled $L0$--$L7$ on the left edge), columns are
  heads $0$--$7$ (labelled $H0$--$H7$ along the top). Each cell
  is the head's average CLS attention on the 64 board squares as
  a heatmap (red, FEN order, $a8$ top-left, $h1$ bottom-right).
  The horizontal bar under each cell shows the head's four region
  masses: board (red) / recent-move (blue) / conditioning
  (green) / other (grey), summing to 1. Side-info-focused heads
  (\texttt{L0.H7}, \texttt{L4.H3}) are visible at a glance.
  Layers $0$--$1$ have moderate board mass and recent-move
  contributions; layers $2$--$4$ are diffuse and conditioning-heavy;
  layers $5$--$7$ progressively re-concentrate on the
  board, with \texttt{L6.H3}'s a8--h1 diagonal and
  \texttt{L7.H4}'s rank-8 band visually striking.}
\label{fig:head_grid}
\end{figure}

Ten of the $64$ (layer, head) units have a visually interpretable
focus, a stable concentration on a specific board area or conditioning 
tokens that holds across all $17$ Phase 3 themes.
The patterns are catalogued in Table~\ref{tab:named_heads}. Two are purely conditioning info
(\texttt{L4.H1} and \texttt{L4.H3}), seven
are purely spatial, and one (\texttt{L0.H7})
is mixed, splitting its mass between the recent-move tokens and a
castled-kingside board region. \texttt{L0.H7} and the two conditioning
heads sit in the diffuse cluster of Table~\ref{tab:head_clusters},
while the seven spatial heads are in the board area cluster. These
ten are the cleanest cases visible in Figure~\ref{fig:head_grid}, not
the only interpretable heads.

\begin{table}[h]
\centering\small
\begin{tabular}{l l p{0.55\linewidth}}
\toprule
Head & Type & Interpretation \\
\midrule
\texttt{L0.H7} & side-info $+$ spatial & Recent-move token tracker -- $41\%$ on positions $0$--$11$, vs $13\%$ uniform. The board portion of its attention concentrates on castled-kingside areas for both colours. \\
\texttt{L2.H3} & spatial & The fianchetto squares (\texttt{b2}, \texttt{g2} for White; \texttt{b7}, \texttt{g7} for Black). \\
\texttt{L2.H4} & spatial & Black castled-king destination squares (\texttt{g8} after O-O; \texttt{c8} after O-O-O). \\
\texttt{L4.H1} & side-info & $40\%$ on the three conditioning tokens (rating, clock, side-to-move), vs $3\%$ uniform. \\
\texttt{L4.H3} & side-info & Highest rating/clock attention mass of any head -- $50\%$ on conditioning tokens, vs $3\%$ uniform. \\
\texttt{L4.H6} & spatial & White's typical piece area, with a particular hotspot on \texttt{g1} (the White king's destination after O-O). \\
\texttt{L5.H5} & spatial & The four central squares (\texttt{d4}, \texttt{e4}, \texttt{d5}, \texttt{e5}) -- the most-contested area. \\
\texttt{L6.H3} & spatial & A striking \texttt{a8}--\texttt{h1} central-diagonal stripe across all themes. \\
\texttt{L7.H4} & spatial & Black's back rank (rank $8$). \\
\texttt{L7.H6} & spatial & White's first two ranks (ranks $1$--$2$). \\
\bottomrule
\end{tabular}
\caption{Named (layer, head) units with visually interpretable
  focus at band $2200$+, identified from
  Figure~\ref{fig:head_grid}. \textbf{Side-info} = the head's
  CLS attention concentrates on the rating/clock/side-to-move or
  recent-move tokens well above the uniform baseline.
  \textbf{Spatial} = the head's CLS-to-board attention has a
  visually distinctive geometry -- a specific square set,
  diagonal, rank, or quadrant -- that is stable across the $17$
  themes in the Phase 3 subset. The two pure side-info heads and the
  mixed \texttt{L0.H7} sit in the diffuse cluster of
  Table~\ref{tab:head_clusters}; the seven pure spatial heads sit in
  the board-focused cluster.}
\label{tab:named_heads}
\end{table}

\begin{figure}[h]
\centering
\includegraphics[page=2,width=0.95\textwidth]{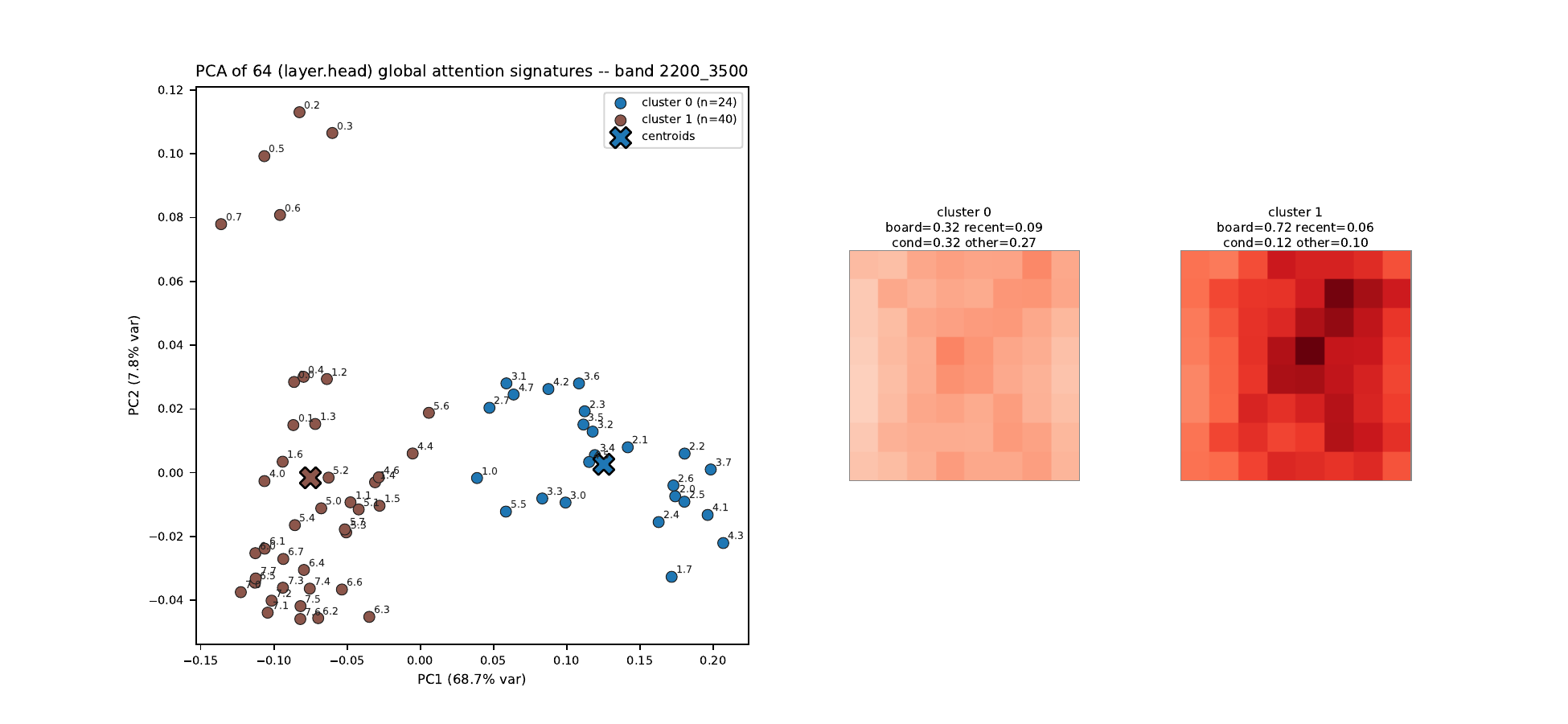}
\caption{Left: PCA scatter of the $64$ (layer, head) global
  attention signatures in $\mathbb{R}^{92}$ at band $2200$+; each
  dot is one (layer, head), labelled $L.H$, coloured by k-means
  cluster ($k\!=\!2$). Right: per-cluster centroid board heatmaps
  (FEN order, $a8$ top-left, $h1$ bottom-right) with the
  centroid's region masses above. Cluster $0$ (blue) is diffuse on
  the board (mass $0.32$); cluster $1$ (red) is board-focused
  (mass $0.72$). The band-1200-1300 panel (not shown) has the same
  cluster shape with a slightly weaker silhouette ($0.375$ vs
  $0.493$).
  Two heads sit far from either centroid along PC2: \texttt{L0.H7}
  (recent-move tracker) and \texttt{L4.H3} (rating/clock attention
  head) -- the named heads in \S\ref{sec:interp}.}
\label{fig:head_clusters}
\end{figure}

Figure~\ref{fig:head_clusters} shows the PCA scatter + per-cluster
centroid heatmaps;
Figure~\ref{fig:head_grid} shows the full $8 \times 8$ grid of
per-head global signatures rendered as $8 \times 8$ board
heatmaps. The head-pruning literature finds that a substantial
fraction of attention heads can be removed at test time without
much loss \cite{michel2019heads}; that $40$ of $64$ heads sit in
the board-focused cluster suggests a similar regime here, though
we have not attempted to prune.

\subsection{Cross-band attention trajectories}
\label{app:trajectories}

To visualize per-band specialization mechanistically, we ran the
same puzzle through three bands (\textit{low} $1200$--$1300$,
\textit{mid} $1700$--$1800$, \textit{high} $2200$+) and laid
the three attention heatmaps side by side. We curate $15$
puzzles by trajectory shape: Type B (low wrong, high right),
Type A (all correct, mass climbs), and Type C (non-monotone). One
exemplar per type is rendered as a Figure below.

In every Figure~\ref{fig:trajectory_B}--\ref{fig:trajectory_C}
panel, the board carries the same legend: the red
overlay is the CLS attention (last layer, head-averaged), and the
dashed black border marks the v1 key squares (the $\{$from, to$\}$
of the target move; identical across all three panels). The
arrow encodes the band's actual played move: \textbf{blue} when
the band predicts correctly (so the arrow runs through the v1
borders), \textbf{red} when the band predicts incorrectly (the
target endpoints are then only marked by the v1 borders, not
duplicated as an arrow). The per-panel title shows the band
label, the model's predicted SAN, the correct/incorrect tag, and
the model's top-$1$ probability mass.

\begin{figure}[h]
\centering
\includegraphics[page=1,width=0.98\textwidth]{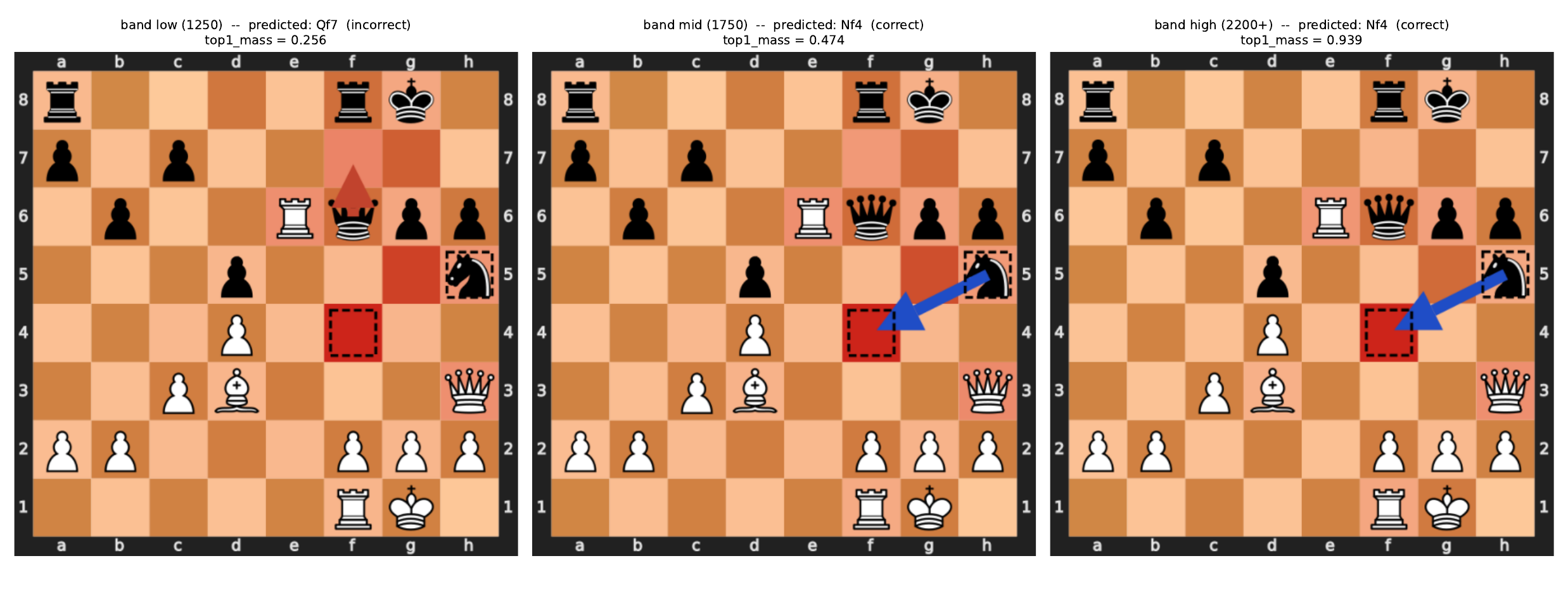}
\caption{\textbf{Type B (qualitative shift)} -- puzzle
  \texttt{r81hY} (rating $1619$, themes \texttt{advantage fork
  long middlegame}, target \texttt{Nf4}). Low: predicts
  \texttt{Qf7} (incorrect), kingside-spread attention, top-$1$
  mass $0.256$. Mid: locks onto \texttt{Nf4} (correct), attention
  on f4, mass $0.474$. High: \texttt{Nf4} (correct), dense
  attention on f4, mass $0.939$. The shift from
  ``queen-on-kingside attention'' to ``knight-on-f4 attention''
  between low and mid bands is the clearest diagnostic pattern of
  per-band specialization in this exemplar.}
\label{fig:trajectory_B}
\end{figure}

\begin{figure}[h]
\centering
\includegraphics[page=12,width=0.98\textwidth]{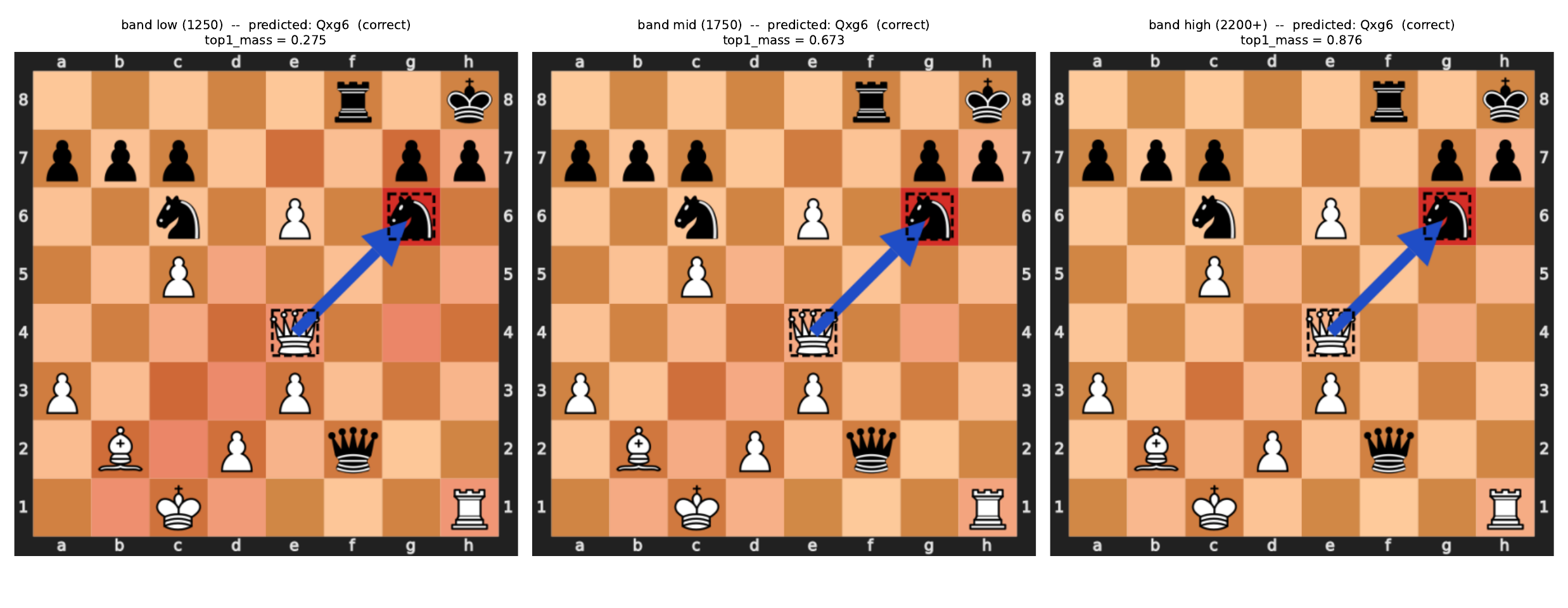}
\caption{\textbf{Type A (sharpening)} -- puzzle \texttt{mQ1iH}
  (rating $2058$, themes \texttt{advantage long middlegame pin},
  target \texttt{Qxg6}). All three bands predict \texttt{Qxg6}
  correctly; top-$1$ mass climbs $0.275 \to 0.673 \to 0.876$.
  Attention is on the same squares (g6 destination, e4 source)
  in all three panels -- just denser at higher bands, illustrating
  the sharpening trajectory while the predicted move is unchanged.}
\label{fig:trajectory_A}
\end{figure}

\begin{figure}[h]
\centering
\includegraphics[page=15,width=0.98\textwidth]{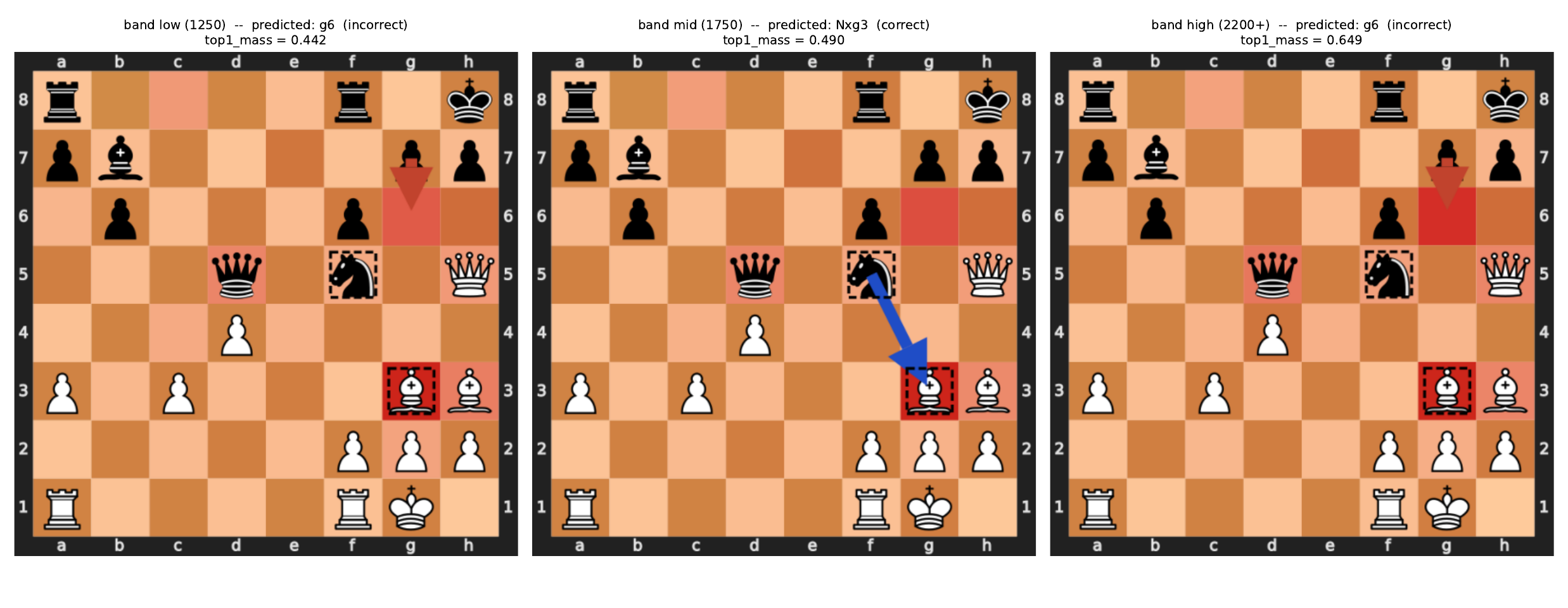}
\caption{\textbf{Type C (non-monotone)} -- puzzle \texttt{t1CJs}
  (rating $1804$, themes \texttt{advantage intermezzo
  kingsideAttack long middlegame}, target \texttt{Ng3}, an
  intermezzo knight check). Low and high bands both pick the
  same wrong move \texttt{g6} (top-$1$ mass $0.44$ low, $0.65$
  high); the mid band finds the correct \texttt{Ng3} (mass
  $0.49$). The $1700$-band model has carved out a window in
  which the intermezzo is visible, while the bracketing bands
  default to a more obvious-looking pawn move. Type C is the
  rarest of the three categories in the candidate pool.}
\label{fig:trajectory_C}
\end{figure}

\subsection{Failure-mode case studies}
\label{app:failures}

Three puzzles from the $800$-puzzle confidently-wrong sub-sample
underpin the typology in Table~\ref{tab:failure_typology}.

\begin{figure}[h]
\centering
\includegraphics[page=1,width=0.98\textwidth]{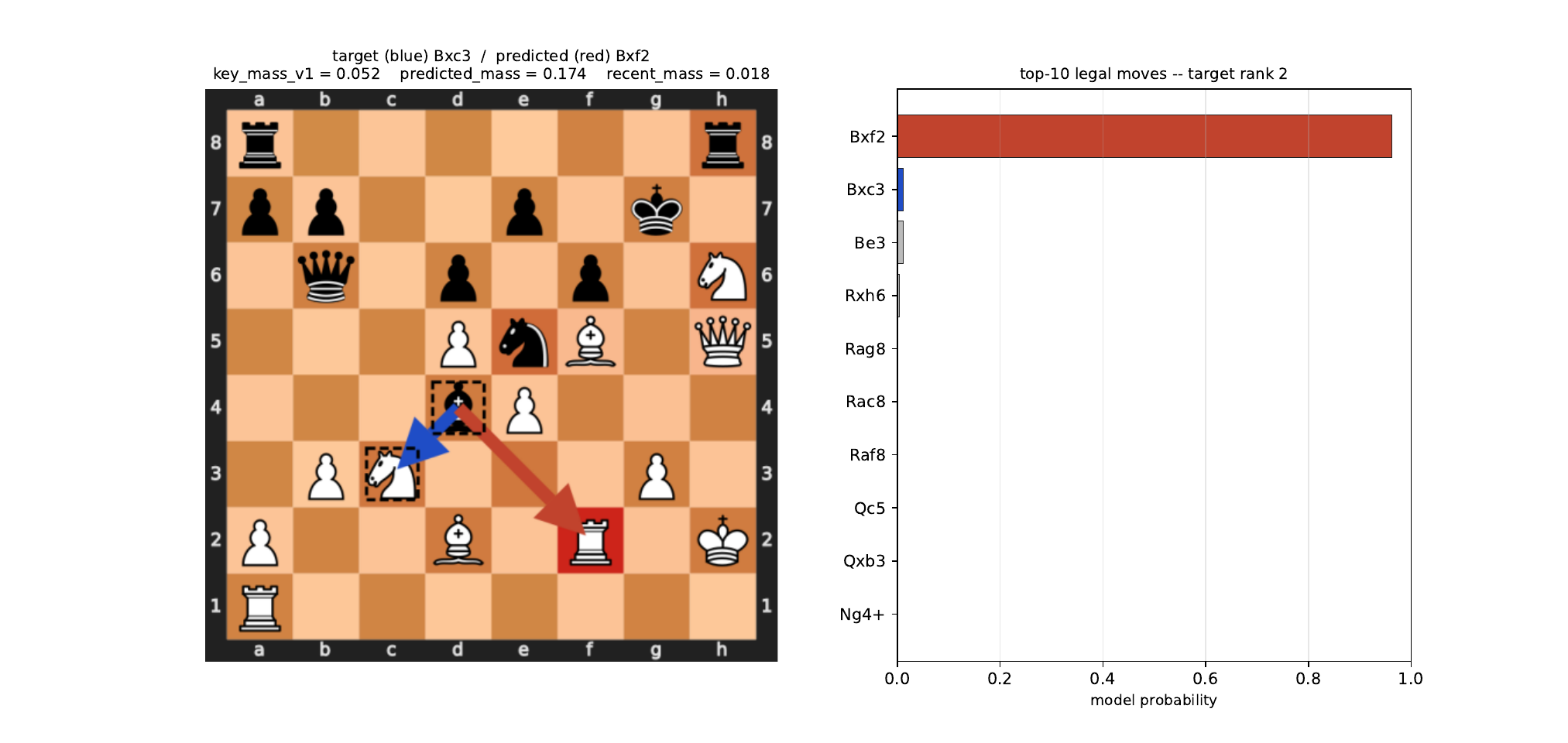}
\caption{\textbf{Type A (attention miss) exemplar} -- puzzle
  \texttt{va7dk} (rating $2325$, themes \texttt{advantage
  middlegame short}). Target: \texttt{Bxc3} (bishop capture
  setting up a discovered attack). Predicted: \texttt{Bxf2}
  (capture the rook on f2) at top-$1$ mass $0.963$; target rank
  $2$ at $<\!0.01$. \texttt{key\_mass}$=\!0.052$ (the model
  essentially never attended to c3); \texttt{predicted\_mass}$=
  \!0.174$. The model committed to the rook capture without
  considering the deeper deflection.}
\label{fig:failure_A}
\end{figure}

\begin{figure}[h]
\centering
\includegraphics[page=4,width=0.98\textwidth]{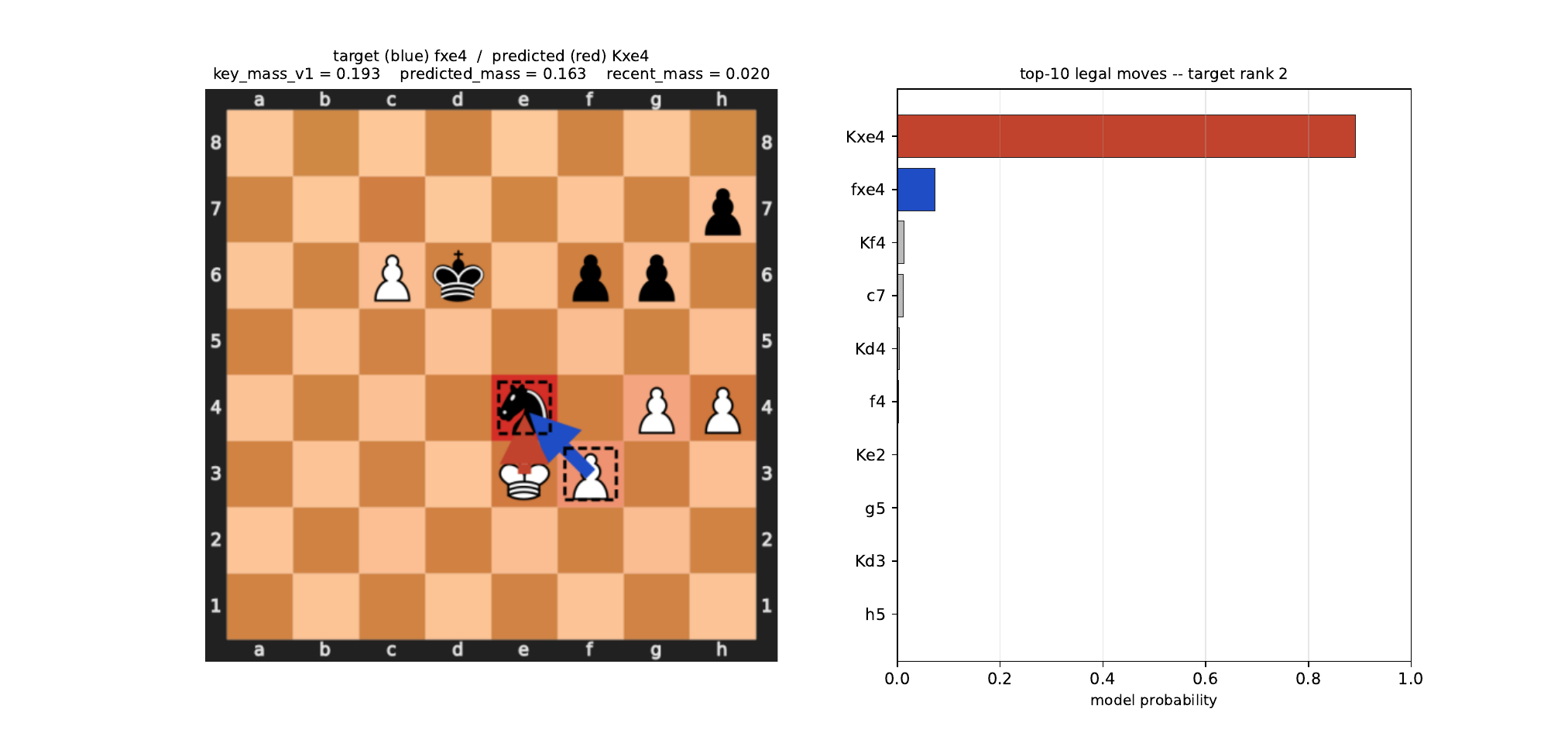}
\caption{\textbf{Type B (right attention, wrong move) exemplar} --
  puzzle \texttt{e7sbs} (rating $2564$, themes \texttt{crushing
  endgame long}). Target: \texttt{fxe4} (pawn captures the knight,
  the correct king-and-pawn endgame technique). Predicted:
  \texttt{Kxe4} (king captures the knight) at top-$1$ mass
  $0.892$; target rank $2$ at $\approx\!0.08$. Both moves land on
  e4 and \texttt{key\_mass}$=\!0.193$ (well above the correct
  mean of $0.140$) -- the model \emph{did} attend to e4. The
  policy head preferred the king over the pawn -- a clean
  ``failure is downstream of attention'' signature.}
\label{fig:failure_B}
\end{figure}

\begin{figure}[h]
\centering
\includegraphics[page=7,width=0.98\textwidth]{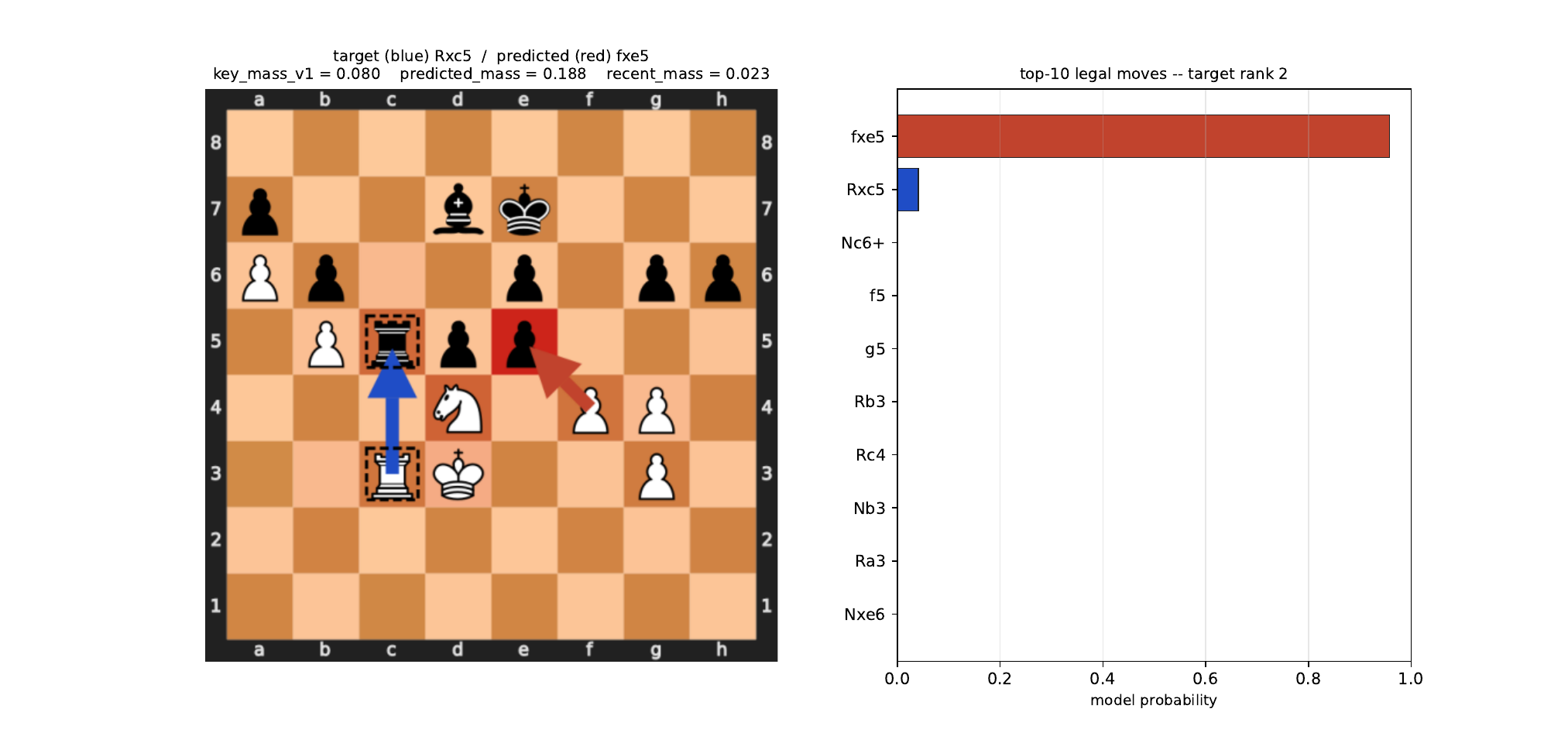}
\caption{\textbf{Type C (recent-mass tail) exemplar} -- puzzle
  \texttt{dJyA4} (rating $2149$, themes \texttt{advancedPawn
  advantage endgame long}). Target: \texttt{Rxc5} (rook captures
  on c5). Predicted: \texttt{fxe5} (pawn captures on e5) at
  top-$1$ mass $0.957$.
  \texttt{recent\_mass}$=\!0.023$ -- just above the pool $p_{90}$
  of $0.022$. The model anchored on the pawn move
  suggested by the local pawn-chain geometry rather than the
  rook capture that the rating-band-specialized solver would
  have found.}
\label{fig:failure_C}
\end{figure}

In every Figure~\ref{fig:failure_A}--\ref{fig:failure_C} the
left panel is the chess.svg board: the red overlay is CLS
attention (last layer, head-averaged); the \textbf{blue} arrow is
the target (correct) move; the \textbf{red} arrow is the model's
predicted (wrong) move; the dashed black border marks the v1 key
squares (the target's $\{$from, to$\}$). The right panel is the
model's top-$10$ legal-move distribution as a horizontal bar
chart, with the target bar coloured blue, the model's top-$1$
bar coloured red, and the remaining bars grey. The per-panel
title above the board reports the target / predicted SAN and the
three diagnostic features (\texttt{key\_mass\_v1},
\texttt{predicted\_mass}, \texttt{recent\_mass}); the bar-chart
title reports the target's rank in the model's softmax.

We interpret high \texttt{key\_mass} as ``the model attended to
the target's endpoints'' rather than as a strict causal claim; the
attention-as-explanation literature
\cite{jain2019attention,wiegreffe2019attention} cautions that
attention weights are a soft proxy for feature importance. Our
typology rests on the \emph{correlation} between attention
pattern and prediction, not on the claim that attention
\emph{causes} the prediction.

\end{document}